\documentclass{article}
\usepackage{microtype}
\usepackage{graphicx}
\usepackage{subfigure}
\usepackage{booktabs}
\usepackage[utf8]{inputenc}
\usepackage{natbib}
\usepackage{hyperref}

\usepackage[accepted]{icml2024}

\usepackage{tabularx}
\usepackage{algorithm}
\usepackage{algorithmic}
\usepackage{amsmath}
\usepackage{amssymb}
\usepackage{mathtools}
\usepackage{amsthm}
\usepackage{multirow}
\usepackage{multicol}
\usepackage{overpic}
\usepackage[capitalize,noabbrev]{cleveref}

\theoremstyle{plain}
\newtheorem{theorem}{Theorem}[section]

\theoremstyle{definition}
\newtheorem{definition}[theorem]{Definition}

\theoremstyle{remark}

\usepackage[textsize=tiny]{todonotes}

\usepackage{pifont}
\newcommand{\xmark}{{\scalebox{0.75}{\ding{55}}}}

\usepackage{enumitem}
\newlist{myitemize}{itemize}{3}
\setlist[myitemize,1]{label=\textbullet,leftmargin=0.1in}

\icmltitlerunning{FuRL: Visual-Language Models as Fuzzy Rewards for Reinforcement Learning}

\begin{document}
\twocolumn[
    \icmltitle{FuRL: Visual-Language Models as Fuzzy Rewards for Reinforcement Learning}
       \icmlsetsymbol{intern}{$\dagger$}
	\begin{icmlauthorlist}
		\icmlauthor{Yuwei Fu}{mcgill,horizon,intern}
        \icmlauthor{Haichao Zhang}{horizon}
        \icmlauthor{Di Wu}{mcgill}
        \icmlauthor{Wei Xu}{horizon}
        \icmlauthor{Benoit Boulet}{mcgill}
	\end{icmlauthorlist}
	\icmlaffiliation{mcgill}{McGill University}
        \icmlaffiliation{horizon}{Horizon Robotics}
	\icmlcorrespondingauthor{Yuwei Fu}{yuwei.fu@mail.mcgill.ca}
 	\icmlcorrespondingauthor{Haichao Zhang}{hczhang1@gmail.com}
	\icmlkeywords{Reinforcement Learning}
	\vskip 0.3in
	]
	\printAffiliationsAndNotice{$^\dagger$ Work done during an internship at Horizon Robotics.} 

\begin{abstract}
    In this work, we investigate how to leverage pre-trained visual-language models (VLM) for online Reinforcement Learning (RL).
    In particular, we focus on sparse reward tasks with pre-defined textual task descriptions.
    We first identify the problem of reward misalignment when applying VLM as a reward in RL tasks.
    To address this issue, we introduce a lightweight fine-tuning method, named Fuzzy VLM reward-aided RL (FuRL), based on reward alignment and relay RL.
    Specifically, we enhance the performance of SAC/DrQ baseline agents on sparse reward tasks by fine-tuning VLM representations and using relay RL to avoid local minima.
    Extensive experiments on the Meta-world benchmark tasks demonstrate the efficacy of the proposed method.
    Code is available at: {\footnotesize\url{https://github.com/fuyw/FuRL}}.
\end{abstract}

\section{Introduction}
    \label{sec:introduction}
    Deep reinforcement learning (RL) has achieved great success in many different domains, including games, robotic control, and graphics~\citep{mnih2015humanlevel, alphago, sac, rubiks_cube, rubiks_cube_openai, dota2, manipulation, amp}.
    However, despite these great achievements, one well-known issue of RL is the large number of environmental interactions required for policy learning~\citep{acer, impala}.

    How to improve the sample efficiency is one of the most important topics in RL~\citep{du2019provably, zhang2020model}.
    A large body of work has been done in the community from different aspects, including better exploration strategy~\citep{icm, gpm}, leveraging in-house behavior data~\citep{parrot}, using transfer learning and (or) meta-learning ~\citep{rakelly2019efficient, mehta2020active, agarwal2023provable, beck2023hypernetworks}, \emph{etc.}

    Recent progress on the large foundation models shows impressive results in many applications~\citep{r3m, liv, vip, rocamonde2023vision, chan2023visionlanguage}.
    These models are useful in the sense that they contain a large amount of common knowledge, which can be used in diverse downstream tasks.
    One promising downstream application is to use the VLM to generate dense rewards for RL tasks with sparse rewards.

    Based on these observations, we investigate how to leverage a pre-trained VLM in online RL.
    The topic of leveraging VLM in the form of reward in RL is an emerging field, with a few recent work on this~\citep{zero_shot_reward, zero_shot_reward_2, languare_reward_modulation, rocamonde2023vision, chan2023visionlanguage}. 
    We follow this line of research and study the issue of reward misalignment when using VLM-based rewards in RL, where inaccurate VLM rewards could trap the agent in local minima.
    To mitigate this issue, we introduce a VLM-representation fine-tuning loss and adopt relay RL~\citep{relayRL} to improve exploration.
    The primary contributions of this work are as follows:
    \vspace{-0.1in}
    
    \begin{itemize}
        \item We investigate some practical challenges of using pre-trained VLM in online RL and highlight the issue of reward misalignment.
        \item We introduce the Fuzzy VLM reward-aided RL (FuRL), a simple yet effective method to address the challenge brought by reward misalignment.
        \item We compare FuRL against different baselines and provide ablation studies to reveal the importance of addressing the fuzzy reward issue.
    \end{itemize}

    \begin{figure*}[tb]
        \centering
        \begin{overpic}[width=2\columnwidth]{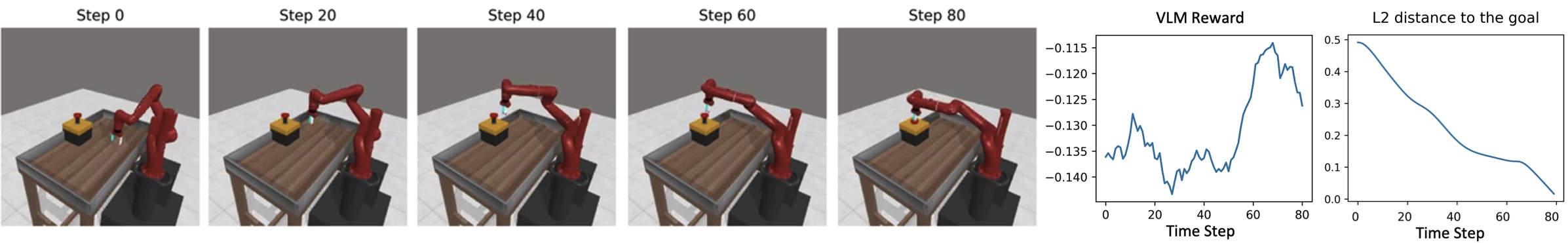}
        \end{overpic}
        \caption{\textbf{Raw VLM reward is sub-optimal to teach RL agents.} In this example, the text instruction $l$ is ``press a button from the top". We plot the cosine similarity-based VLM reward with language embedding $\Phi_L(l)$ and image embedding $\Phi_I(o_{t})$ and also show the distance between the end-effector and the goal. It can be observed that the cosine similarity between $\Phi_L(l)$ and $\Phi_I(o_{t})$ can reflect some aspects of the task but is not always well aligned with the task progress, reflecting the fuzzy aspects of the VLM reward.}
        \label{fig:vlm_repr}
    \end{figure*}
 
\section{Background}
    \subsection{Markov Decision Process (MDP) and RL}
        An MDP~\citep{sutton_book} is commonly defined by a tuple $(\mathcal{S, A}, P, r, \gamma)$, where $\mathcal{S}$, $\mathcal{A}$ denote the state space and action space. 
        $P: \mathcal {S \times A} \to \Delta(\mathcal S)$ denotes the transition probability between states. $r: \mathcal{S\times A} \!\rightarrow\!\mathbb{R}$ is the reward function. $\gamma\!\in\![0, 1]$ denotes the discount factor. 
        In the standard RL formulation, our goal is to learn a policy that maximizes the expected accumulated discounted return. In practice, the policy is typically modelled using a neural network $\pi_{\theta}(a_t|s_t)$ with learnable parameters $\theta$~\citep{mnih2013playing}, taking observation $s_t$ as input and generating the action from policy $\pi_\theta(a_t|s_t)$ for each time step $t$.

    \subsection{Vision Language Models (VLM)}
        Vision Language Models have advanced rapidly in the past few years~\citep{clip, flamingo, liv}.
        One representative work is CLIP~\citep{clip}, which trains the VLM by aligning image and text embedding in the latent space. CLIP has been shown to be effective in downstream tasks such as classification and
        also shows zero-shot transfer ability.
        While CLIP is a generic model motivated by vision tasks, there is also recent work
        on specially designed VLM for RL tasks~\cite{liv}.
        VLM is an attractive type of models for aiding RL from different perspectives, such as reward shaping~\citep{zero_shot_reward, zero_shot_reward_2, languare_reward_modulation, rocamonde2023vision, chan2023visionlanguage, foundation_model_reward, clipmotion, motif}, task specification and success detection~\cite{VLM_success_detector} and representation~\cite{chen2023visionlanguage}.
   
\section{Method}
    \label{sec:method}
    In this section, we first revisit some existing work on utilizing VLM as a reward model in RL and pinpoint the challenges therein.
    We then introduce the main idea and formulation of the proposed method.

    \begin{figure*}[thb]
        \centering
        \begin{overpic}[width=2\columnwidth]{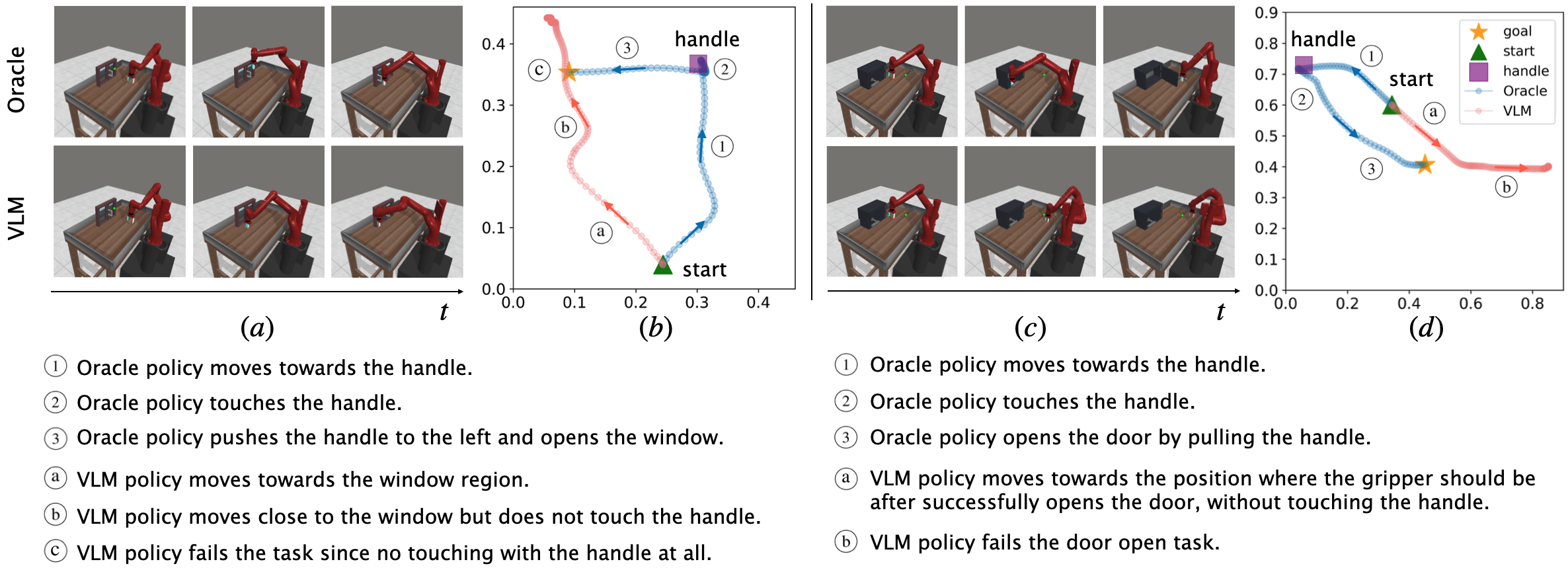} 
        \end{overpic}
        % \vspace{-0.18in}
        \caption{\textbf{Fuzzy VLM reward effect.} Visualization of end-effector trajectory in terms of $(x, y)$ positions. \emph{Oracle} denotes an expert policy. \emph{VLM} denotes the policy trained using sparse task reward together with VLM reward.}
        \vspace{-0.15in}
        \label{fig:challenges}
    \end{figure*}
        
    \subsection{VLM as Rewards Revisited}
        \label{sec:vlm_as_reward}
        Leveraging VLM as a source of reward in RL is a popular and active emerging trend~\citep{zero_shot_reward, zero_shot_reward_2, languare_reward_modulation, rocamonde2023vision, chan2023visionlanguage, foundation_model_reward, clipmotion, motif, lift}, either as a way of reward-based task specification~\citep{zero_shot_reward, zero_shot_reward_2, languare_reward_modulation}, or generating VLM-based reward as an additional source of supervision apart from the original task reward for RL~\citep{ rocamonde2023vision, chan2023visionlanguage, foundation_model_reward, clipmotion, motif}.

        Given an observation $s_t$ received at timestep $t$, the RL agent 
        generates an action $a_t\sim \pi_{\theta}(a_t|s_t)$ and 
        receives a sparse task reward $r^{\rm task}_t$ after $a_t$ is executed.
        $r^{\rm task}_t$ is typically defined as $r^{\rm task}_t=\delta_{\rm success}$, meaning a reward of 1 is received only upon task success and otherwise the reward is 0.

        This is a type of task setting commonly encountered in practice. The sparse reward 
        makes the RL training more challenging.
        \cite{rocamonde2023vision, chan2023visionlanguage} propose to use \emph{VLM as reward}, \emph{i.e.},
        augmenting the sparse task reward with another VLM reward $r^{\rm VLM}_t$:
        \begin{equation}
            \label{eq:vlm_reward}
            r_t = r^{\rm task}_t + \rho \cdot r^{\rm VLM}_t,
        \end{equation}
        where $\rho$ is a scalar weight parameter for balancing the VLM reward with the task reward.
    
        Simply, these methods~\cite{rocamonde2023vision, chan2023visionlanguage} add the CLIP reward, \emph{i.e.} the cosine similarity between the language goal with an image of the latest state:
        \begin{equation}
            \label{eq:r_vlm}
            r^{\rm VLM}_t \triangleq r^{\rm CLIP}_t = \frac{\langle \Phi_L(l), \Phi_I(o_t) \rangle}{\Vert\Phi_L(l)\Vert\cdot \Vert\Phi_I(o_{t})\Vert},
        \end{equation}
        to the sparse task reward.
        $o_{t}$ is the image observation received at step $t$.
        $l$ is the language-based task instruction issued at the beginning of the episode.
        $\Phi_L$ and $\Phi_I$ denote the language embedding network and image embedding network of the pre-trained CLIP model~\cite{clip}.
    
        Another related set of work is using VLM as a success detector~\cite{VLM_success_detector}, \emph{i.e.}, as a sparse task reward.
        Apart from RL-based policy training, some recent work also proposed to use VLM-based reward for model-based planning~\citep{liv}.
        In this case, the task reward is omitted (\emph{i.e.} $r_t = r^{\rm VLM}_t$) and MPC type of online planning methods are used to obtain the next action by maximizing future return.

        In this work, we follow the line of research on VLM-as-reward.
        Same as~\citet{rocamonde2023vision,chan2023visionlanguage}, we focus on sparse-reward tasks, 
        and assume the access to the task instruction $l$ (Table~\ref{tab:language_goal}) and a goal image $o_g$ at the beginning of the episode.
        Without loss of generality, we first present results using state-based observations as policy input to disentangle the impacts of feature learning.
        We then increase the complexity by using pixel-based observations as policy input.
        For computing the VLM feature, a visual image is provided at each time step.

        \begin{figure*}[tb]
            \centering
            \begin{overpic}[width=2\columnwidth]{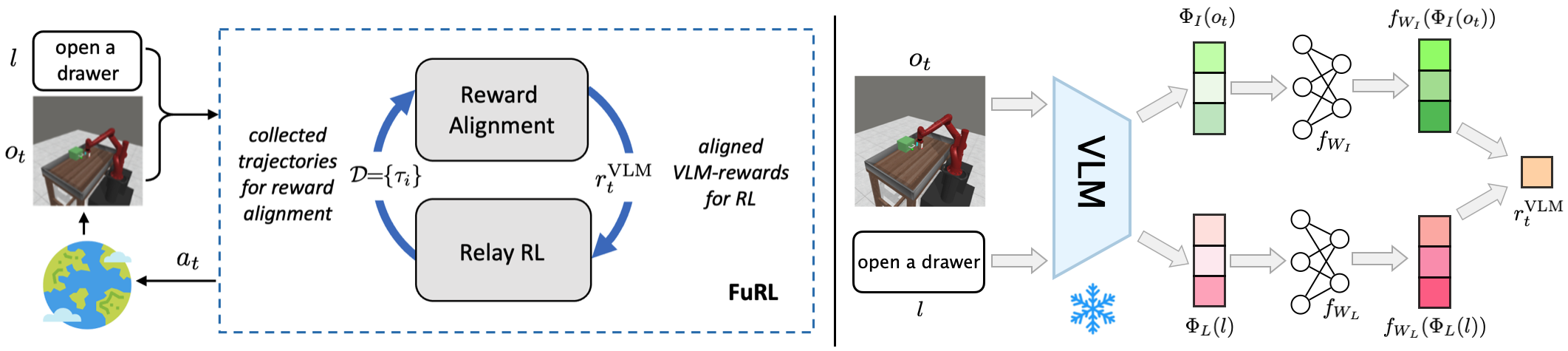}
            \end{overpic}
            % \vspace{-0.1in}
            \caption{\textbf{Illustration of the proposed method:} (left) the overall pipeline of FuRL. (right)
            FuRL freezes the pre-trained VLM and only fine-tunes two MLP-based projection heads  $f_{W_L}$, $f_{W_I}$.}
            % \vspace{-0.1in}
            \label{fig:framework}
        \end{figure*}

    \subsection{VLM as \emph{Fuzzy} Rewards}
        \label{sec:challenge}
        Many previous methods on \emph{VLM-as-rewards}~\citep{rocamonde2023vision, chan2023visionlanguage, zero_shot_reward,liv} have a shared assumption that VLM-based rewards are accurate in order to achieve good policy optimization 
        to avoid undesired solutions.

        In this work, we demonstrate that \emph{zero-shot VLM rewards are fuzzy}:  meaningful in capturing the coarse semantics but inaccurate in characterizing some details.
        Therefore, this fuzziness in the VLM-based rewards could potentially mislead the policy optimization in the \emph{VLM-as-reward} framework~\citep{chan2023visionlanguage,zero_shot_reward,liv}.
        While the degree of the reward's fuzziness can be reduced by 
        changing different aspects of the VLM model, such as increasing its capacity, the reward's fuzziness is not likely to be eliminated due to the zero-shot nature of the VLM-as-reward framework.
        We carry out two sets of studies from complementary aspects to illustrate this point.
        
        {\textbf{Rewards along Expert Trajectory.}}
        Figure~\ref{fig:vlm_repr} shows the VLM reward curve for an expert trajectory, where we compute the VLM reward $r^{\rm VLM}_t$ using Eqn.~\ref{eq:r_vlm}.
        In the ideal case, the reward curve should be aligned with the expert's progress, \emph{i.e.},
        higher reward when the state is closer to the task completion.
        However, as can be observed in Figure~\ref{fig:vlm_repr}, the reward curve can reflect some aspects of task progress but is not well aligned with the task progress,  reflecting the fuzzy aspect of the VLM reward.

        {\textbf{VLM Reward Only Policy Behavior.}}
        We also trained a VLM policy only with $r^{\rm VLM}$.
        Figure~\ref{fig:challenges} illustrates the end-effector (gripper)  trajectories of the robot arm in the \textit{window-close} and \textit{door-open} tasks from the Meta-world environment~\citep{metaworld}.
        We compared the trajectories generated by an oracle policy with those of a VLM policy.
        As depicted in Figure~\ref{fig:challenges} (a-b) and Figure~\ref{fig:challenges} (c-d), we can observe that the gripper is close to the window in the \textit{window-close} task and the gripper is far away from the door in the \textit{door-open} task at the last step, also exemplifying 
        the effect of fuzzy VLM reward on policy behavior.

        These two set of case studies show that VLM reward sometimes could provide meaningful information, \textit{i.e.}, identifying the window in \textit{window-close} task.
        However, when pre-trained VLM representations fail to capture crucial information in the target RL tasks, inaccurate VLM rewards can hinder efficient exploration.
        For instance, as seen in Figure~\ref{fig:challenges}(c-d), the robot arm got stuck at the right corner in the \textit{door-open} task and failed to collect any successful trajectories during the training.
        Such inaccurate VLM rewards are mainly due to the domain shift between VLM's training dataset and the downstream target RL task~\citep{sankaranarayanan2018learning, zhang2021adaptive}.
        
        All these results indicate that the VLM-based rewards are fuzzy, \emph{i.e.}, meaningful in some cases but could also be misleading due to their inaccuracy.
        This fuzzy reward issue has caught some attentions very recently~\citep{zero_shot_reward, languare_reward_modulation, rocamonde2023vision}.
        \citet{languare_reward_modulation} mitigate this issue by using VLM-based reward for behavior pre-training only.
        \citet{zero_shot_reward} retrain the VLM model by using a specially tailored dataset. 
        We instead focus on how to leverage a pre-trained VLM model in  online RL and strategies to mitigate the challenges therein, as presented in the sequel.

    \subsection{FuRL: Fuzzy VLM rewards-aided RL}
        In this subsection, we introduce the \textit{Fuzzy VLM rewards-aided RL} (FuRL), a framework that utilizes VLM rewards to facilitate learning in sparse reward tasks while addressing the inherent fuzziness of these rewards through two mechanisms: (1) reward alignment and (2) relay RL, as depicted in Figure~\ref{fig:framework} (left).
        These two components interact with each other in terms of exploration and learning in FuRL:
        {\textbf{\emph{(i)}}}~{\textbf{Reward Alignment}}: which fine-tunes VLM representations (generated embeddings) in a lightweight form to improve the VLM rewards, which helps exploration and policy learning;
        {\textbf{\emph{(ii)}}}~{\textbf{Relay RL}}: which helps to escape the local minima due to the fuzzy VLM rewards during exploration, and it also helps to collect more diverse data to improve the reward alignment and policy learning.
        We will detail these components in the following subsections respectively.

        \subsubsection{Reward Alignment}
            It is natural to understand that a VLM-based reward function, as an instance of learning-based
            reward function, is hard to be accurate under all kinds of input variations.
            Reward inaccuracy is undesirable since it could be misleading to the policy learning~\citep{skalse2022defining}.
                
            With cosine similarity, inaccurate VLM rewards as defined in Eqn.~\ref{eq:r_vlm}  can be attributed to the misalignment between image and text embedding from pre-trained VLM representations.
            To address this  issue, we introduced a lightweight alignment method as illustrated in Figure~\ref{fig:framework} (right).
            In particular, we freeze the pre-trained VLM and only append two small learnable networks $f_{W_L}$ and $f_{W_I}$ to VLM's text embedding and image embedding, respectively.
            In our experiments, $f_{W_L}$ and $f_{W_I}$ are two simple two-layer MLPs~\citep{tolstikhin2021mlp}.
            Therefore, compared with fine-tuning the whole VLM model, the number of parameters to be learned in our method is much smaller.
    
            We define the VLM reward via the cosine-similarity following \citep{rocamonde2023vision, chan2023visionlanguage} but with our projected image embedding $f_{W_I}(\Phi_I(o_{t}))$ and the projected text embedding $f_{W_L}(\Phi_L(l))$:
            \begin{equation}
                \label{eq:projected_reward}
                r^{\text{VLM}}_t \triangleq  \frac{\langle f_{W_L}(\Phi_L(l)), f_{W_I}(\Phi_I(o_{t}))\rangle}{\Vert f_{W_L}(\Phi_L(l))\Vert\cdot \Vert f_{W_I}(\Phi_I(o_{t}))\Vert}.
            \end{equation}
    
            Next, we introduce the following definition:
            \begin{definition} (\emph{Reward Alignment})
            Given a target task $\mathcal{T}$, and a reward function $r$,
                we define the process of adjusting the initially inaccurate reward function $r$ to be more accurate in characterizing the target task $\mathcal{T}$ as reward alignment.
            \end{definition}
            Since the sparse reward function $r^{\rm task}$ in the MDP characterizes the target task to a great extent, we will leverage the information from $r^{\rm task}$ for reward alignment, \emph{i.e.}, optimizing the projection network $f_{W_L}$ and $f_{W_I}$.
            We denote the samples from the successful trajectories $\tau^p$ as \textit{positive} samples $o^p$, and the samples from the unsuccessful trajectories  $\tau^n$ as \textit{negative} samples $o^n$.
            We propose the following loss for reward alignment:
            \begin{equation}\label{eq:reward_alignment}
                \mathcal L= \underbrace{\displaystyle \mathop{\mathbb{E}}_{\{o^p\in \tau^p, o^n\in \tau^n\}} \mathcal{\ell}_{\delta}(o^p, o^n)}_{\mathcal L_{\text{pos-neg}}} + \underbrace{\displaystyle \mathop{\mathbb{E}}_{\{o^p_{i-k}, o^p_i\in \tau^p\}} \mathcal{\ell}_{\delta}(o_i^p, o_{i-k}^p)}_{\mathcal L_{\text{pos-pos}}},
            \end{equation}
            where  $\mathcal{\ell}_\delta(o_p, o_n)\triangleq \max(0, r^{\rm VLM}(o_n) - r^{\rm VLM}(o_p) + \delta)$ is a ranking loss with a margin of $\delta \in \mathbb{R}^+$, generating a loss if $r^{\rm VLM}(o_n) + \delta$ is larger than $r^{\rm VLM}(o_p)$.
            $\mathcal L_{\text{pos-neg}}$ learns to generate a higher VLM reward for a positive sample than that of a negative sample.
            $\mathcal L_{\text{pos-pos}}$ learns to rank two samples from the same successful trajectory,
            giving samples later in time a higher rank (larger score) since it is closer to the task success.
            Here, $k$ is a window size parameter.
            \begin{figure}[tb]
                \centering
                \begin{overpic}[width=.95\columnwidth]{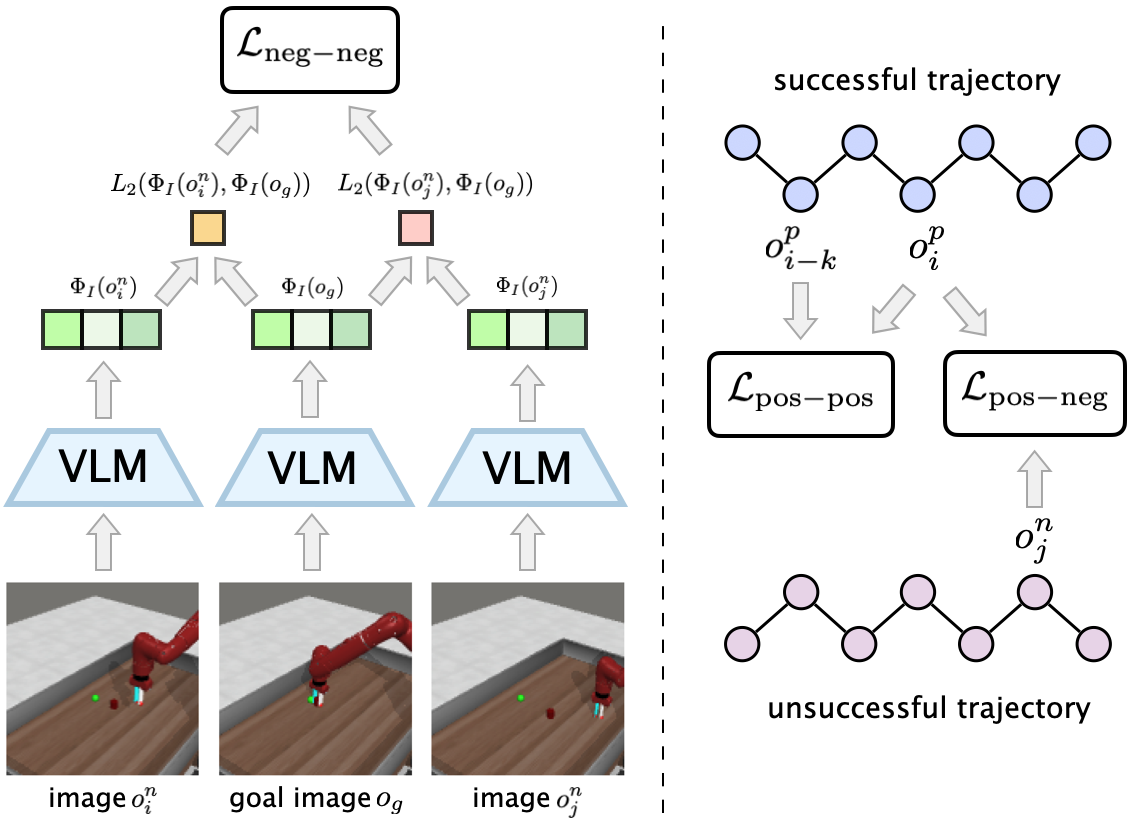}
                \end{overpic}
                % \vspace{-0.1in}
                \caption{\textbf{Contrastive learning loss}: (left) without any successful trajectories, we can use L2 distance w.r.t. an goal image to rank the goodness of two negative samples; (right) when we collected some successful trajectories, the contrastive loss learns to distinguish samples from both of the successful and unsuccessful trajectories.}
                % \vspace{-0.1in}
                \label{fig:contrastive_loss}
            \end{figure}

            \begin{figure*}[tb]
                \centering
                \begin{overpic}[width=1.95\columnwidth]{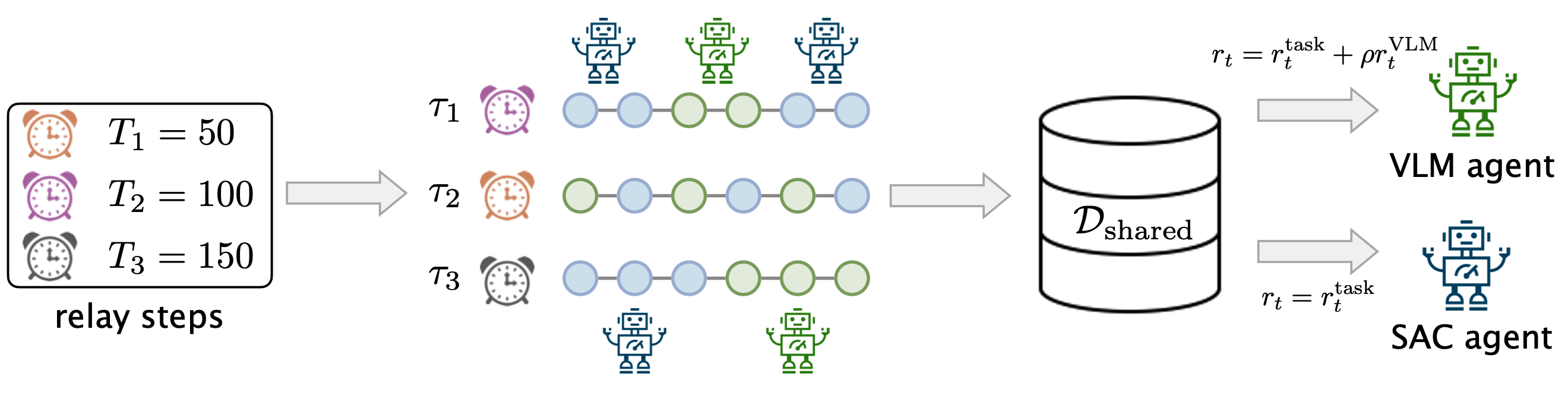}
                \end{overpic}
                % \vspace{-0.2in}
                \caption{\textbf{Illustration of the relay RL based exploration:} at the beginning of an episode $\tau_i$, we randomly select a relay step $T_i$. We iteratively unroll a VLM agent and a SAC agent for $T_i$ steps and save the collected samples in a shared buffer. We turn off the relay exploration when we collected 2500 positive samples from the successful trajectories.}
                \vspace{-0.05in}
                \label{fig:relay}
            \end{figure*}

            Moreover, since successful trajectories are unavailable in the beginning of the training,  an optional step can be used when an additional goal image $o_g$ is available before encountering any successful trajectories, \emph{i.e.}, learning from all \textit{negative} samples with zero task rewards:
            \begin{equation}
                \label{eq:loss_neg}
                \mathcal L_{\text{neg-neg}}= \displaystyle \mathop{\mathbb{E}}_{\substack {\{o^n_i, o^n_j\in \tau^n} \\  {L_2(o^n_i, o_g)<L_2(o^n_j, o_g)-\delta'\}}}
                \mathcal{\ell}_{\delta}(o^n_i, o^n_j),
            \end{equation}
            with $L_2(o, o_g)\triangleq \Vert \Phi_I(o)- \Phi_I(o_g) \Vert_2$.
            This essentially uses the distance w.r.t. the goal image $o_g$ in the image embedding space to rank the goodness of samples, ranking samples with smaller distance higher (Figure~\ref{fig:contrastive_loss}).
            The main purpose of Eqn.~\ref{eq:loss_neg} is to accelerate the learning to find the first successful trajectory earlier.
            In practice, we can also replace $\mathcal L_{\text{neg-neg}}$ by using parallel agents or exploration intrinsic reward to search for the first successful trajectory. We leave the exploration of this as a future work.

            \begin{algorithm}[h]
               \caption{Fuzzy VLM rewards aided RL (FuRL)}
               \label{alg:code}
                \begin{algorithmic}
                   \STATE {\bfseries Input:} pre-trained VLM $\Phi_I$ and $\Phi_L$, goal image $o_g$, task language goal $l$, relay steps $T_s = [T_1, \cdots, T_n]$, shared replay buffer $\mathcal D_{\text{shared}}$, total trajectory number $N$.
                   \STATE {\bfseries Output:} trained VLM agent $\pi_{\rm VLM}$.
                   \STATE {\bfseries Initialize:} image projection head $f_{W_I}$, language projection head $f_{W_L}$, VLM agent $\pi_{\rm VLM}$, SAC agent $\pi_{\rm SAC}$.
                    \FOR{$i=1$ {\bfseries to } $N$}
                        \IF{Collected positive samples}
                            \STATE Unroll VLM policy $\pi_{\rm VLM}$.
                            \STATE Update $f_{W_I}$ and $f_{W_L}$ using Eqn.~\ref{eq:reward_alignment}.
                            \STATE Update $\pi_{\rm VLM}$ using $r^{\rm task} + \rho r^{\rm VLM}$.
                        \ELSE
                            \STATE Sample a replay step $T_i$ from $T_s$.
                            \STATE Iteratively unroll $\pi_{\rm VLM}$ and $\pi_{\rm SAC}$ for $T_i$ steps.
                            \STATE Update $f_{W_I}$ and $f_{W_L}$ using Eqn.~\ref{eq:loss_neg}.
                            \STATE Update $\pi_{\rm SAC}$ using $r^{\rm task}$.
                            \STATE Update $\pi_{\rm VLM}$ using $r^{\rm task} + \rho r^{\rm VLM}$.
                        \ENDIF
                    \ENDFOR
                \end{algorithmic}
            \end{algorithm}
            
        \subsubsection{Relay RL}
            As previously mentioned, one notable challenge in VLM reward alignment is how to find the first successful trajectory earlier.
            When the current VLM policy is trapped in local minima due to the inaccurate VLM rewards, as shown in Figure~\ref{fig:challenges}, it is likely that the agent fails to collect any successful trajectories.
            Under such circumstances, Eqn.~\ref{eq:reward_alignment} is never triggered as we have no positive samples.

            Given this observation, we introduce a simple exploration strategy based on the relay RL~\citep{gupta2020relay, relayRL} to mitigate this representative issue caused by fuzzy VLM reward.
            More specifically, we maintained an extra SAC agent $\pi_{\rm SAC}$ besides the current VLM agent $\pi_{\rm VLM}$.
            At the beginning of an episode $\tau_i$, we first randomly select a relay step $T_i$ from some pre-defined values or a specified range.
            We then iteratively unroll $\pi_{\rm SAC}$ and $\pi_{\rm VLM}$ for $T_i$ steps until the end of the trajectory, as shown in Figure~\ref{fig:relay}.
            The collected samples are added to a shared buffer, which we use to train $\pi_{\rm VLM}$ with $r^{\rm task}_t + \rho r^{\rm VLM}_t$ and  $\pi_{\rm SAC}$ with $r^{\rm task}_t$.
            The evaluation is done on the VLM agent.

            The motivation of the relay RL is to let the SAC agent help to escape the local minima once the VLM agent gets stuck.
            On the other hand, relay RL also helps to increase the data diversity by starting $\pi_{\rm VLM}$ and $\pi_{\rm SAC}$ from different initial states.
            Generally, starting with $\pi_{\rm VLM}$ forms a curriculum learning for the SAC agent as in Jump-start RL~\citep{uchendu2023jump}.
            Once we have collected some successful trajectories, we can turn off the relay RL and focus on collecting samples with the VLM policy $\pi_{\rm VLM}$.
            The pseudo-code of FuRL is summarized in the Algorithm ~\ref{alg:code}.

\section{Related Work}

    \begin{table*}[htb]
        \caption{\textbf{Experiment results on the MT10 benchmark with sparse reward and fixed goal.} We report the average success rate $P$ ($\%$) in the evaluation at the last timestep across 5 random seeds after training.}
        \label{tab:mt10}
        \footnotesize
        \begin{center}
        \resizebox{17cm}{!}{
            \begin{tabular}{lccccccccc}
            \toprule
            \multirow{3}{*}{Environment} & & SAC & VLM-RMs & VLM-RMs-GB & LIV & LIV-Proj & Relay  & FuRL w/o goal-image & {FuRL}\\  \cmidrule{2-10}
             & $r^{\rm VLM}$ &  \xmark &  \checkmark & \checkmark& \checkmark & \checkmark & \checkmark & \checkmark & \checkmark \\
            & $r^{\rm task}$ & \checkmark & \xmark &  \xmark &  \checkmark & \checkmark & \checkmark & \checkmark & \checkmark \\
            \midrule
            \multicolumn{2}{l}{button-press-topdown-v2} & 0 & 0 & 0 & 0 & 0 & 60 & 80 & 100 \\
            \multicolumn{2}{l}{door-open-v2}            & 50 & 0 & 0 & 0 & 0 & 80  & 100 & 100 \\
            \multicolumn{2}{l}{drawer-close-v2}         & 100 & 0 & 0 & 100 & 100 & 100 & 100
 & 100 \\
            \multicolumn{2}{l}{drawer-open-v2}          & 20 & 0 & 0 & 0 & 0 & 40 & 80 & 80  \\
            \multicolumn{2}{l}{peg-insert-side-v2}      & 0 & 0 & 0 & 0 & 0 & 0  & 0 & 0 \\
            \multicolumn{2}{l}{pick-place-v2}           & 0 & 0 & 0 & 0 & 0 & 0 & 0 & 0   \\
            \multicolumn{2}{l}{push-v2}                 & 0 & 0 & 0  & 0 & 0 & 0 & 40 & 80  \\
            \multicolumn{2}{l}{reach-v2}                & 60 & 0 & 0 & 80 & 80 & 100 & 100 & 100 \\
            \multicolumn{2}{l}{window-close-v2}         & 60 & 0 & 0 & 60 & 40 & 80 & 100 & 100 \\
            \multicolumn{2}{l}{window-open-v2}          & 80 & 0 & 0 & 40 & 20 & 80 & 100 & 100 \\
            \midrule 
            \multicolumn{2}{l}{average} & 37.0 & 0.0 & 0.0 & 28.0 & 24.0 & 54.0  & 70.0 & \textbf{76.0}\\
            \bottomrule
            \end{tabular}
            }
        \end{center}
        \vspace{-0.1in}
    \end{table*}

    \subsection{RL with VLM}
        As one specific type of foundation models, VLM connects language with visual signals and has been playing an important role in the fields that involves both modalities such as 
        visual question answering~\citep{agrawal2016vqa, Das_2018_CVPR, zhang2023visionlanguage}. 
        VLM has been used in RL in various ways. It has been used as a reward function~\citep{zero_shot_reward, zero_shot_reward_2, languare_reward_modulation, rocamonde2023vision, chan2023visionlanguage, foundation_model_reward, clipmotion, motif, lift}, as revisited in detail in Section~\ref{sec:vlm_as_reward}. \citet{sontakke2023roboclip} also used VLM for reward computation but requires additional expert demonstrations.
        \citet{internet_vlm} used a generative VLM for hindsight relabeling-based data augmentation to improve dataset diversity.

        Apart from this, VLM has also been used in other ways such as a promptable representation learner~\citep{chen2023visionlanguage}.
        In this work, we focus on the VLM-as-reward setting.
        In terms of training, VLM can be trained in a general way, without being tailored to the downstream tasks.
        Recently, robotics/RL oriented VLM are emerging~\citep{pgvlm2023,liv}.
        The backbone VLM of this work is base on one of such a model~\citep{liv}.

    \vspace{-0.05in}
    \subsection{RL with Foundation Models}
        Our work falls within the broader category of leveraging foundation models in RL, which is an active field with many exciting advances~\citep{foundation_model_reward, lift, voyager, foundation_models_for_robot}.
        Apart from VLM, other forms of foundation models such as large language models (LLM)
        have also been used in many different ways, including planning~\citep{lm_zeroshot_planner, inner_monologue}, task decomposing with grounding~\citep{saycan, grounded_decoding, bootstrap_skill},
        generating code as policy/skill~\citep{liang2023code},
        reward design~\citep{lm_reward_syn, ma2023eureka} \emph{etc.}.
        In this paper, we focus on a complementary perspective by highlighting the potential issues of using a pre-trained VLM in RL and proposing practical remedies.

% \vspace{-0.05in}
\section{Experiments}
% \vspace{-0.05in}
\label{sec:experiments}

    In this section, we focus on the following questions:
    (1) How does the proposed FuRL perform compared to other baselines?
    (2) Is FuRL effective with pixel-based observations?
    (3) Can FuRL generalize to other VLM backbone models?
    (4) Are both the reward alignment and relay RL components useful?
    (5) What is the influence of the VLM reward weight parameter $\rho$?

    \subsection{Baselines}
        To validate the efficacy of the proposed method, we compare the proposed FuRL to the following baselines:
        \begin{enumerate}
            \item SAC: a state-based SAC agent~\citep{sac} using the sparse binary task reward $r^{task}_t$.
            \item VLM-RMs~\citep{rocamonde2023vision}: a recent baseline which only uses the cosine similarity based VLM reward without the task reward.
            \item VLM-RMs-GB~\citep{rocamonde2023vision}: a variant of VLM-RMs which adds a goal-baseline regularization.
            \item LIV~\citep{liv}: a state-based SAC agent trained with task reward and dense LIV reward as in Eqn.~\ref{eq:vlm_reward}.
            \item LIV-Proj: similar to LIV baseline but with the LIV reward computed as in Eqn.~\ref{eq:projected_reward}  using randomly initialized and fixed $f_{W_L}$ and $f_{W_I}$.
            \item Relay~\citep{relayRL}: a simplified version of FuRL where we incorporated relay into LIV baseline.
            \item FuRL-without-image: a variant of FuRL that does not use the goal image and starts to fine-tune only after collecting the first successful trajectory.
        \end{enumerate}
    
        \begin{figure*}[t]
            \centering
            \begin{overpic}[width=2.0\columnwidth]{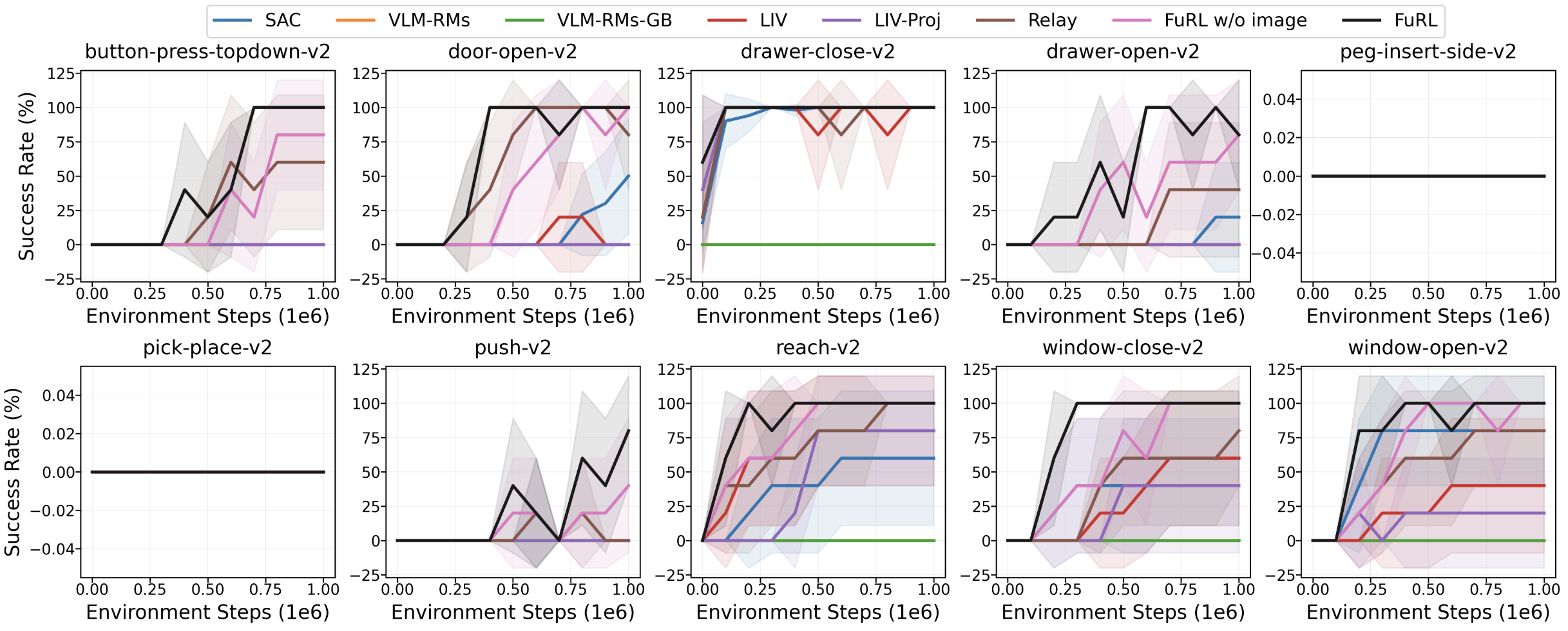}
            \end{overpic}
            % \vspace{-0.1in}
            \caption{\textbf{Evaluation curves on the Sparse Meta-world tasks}: FuRL generally outperforms the other baselines.}
            % \vspace{-0.2in}
            \label{fig:mt10}
        \end{figure*}

    \subsection{Experiment Settings}
         We use ten robotics tasks from the Meta-world MT10 environment~\citep{metaworld} with state-based observations and sparse rewards (referred to as \emph{Sparse Meta-world Tasks}).
         In each task, the RL agent only receives reward 1 when it reaches the goal and otherwise the reward is 0.
         We use SAC~\citep{sac} as the backbone RL agent, and the total environmental step is 1e6.
         We use the Adam optimizer with a learning rate of 0.0001.
         The VLM reward weight $\rho$ is 0.05.
         For the VLM model, we use the pre-trained LIV~\citep{liv} from the official implementation.
         We provide more detailed information in the Appendix~\ref{appendix:exp_setup}.
         More results and resources are available on the project page.~\footnote{\scriptsize{\url{https://sites.google.com/site/hczhang1/projects/furl}}}

    \subsection{Results on Sparse MT10}
        \label{subsec:mt10_}

        We first validate the effectiveness of FuRL on the fixed-goal MT10 benchmark~\citep{metaworld}.
        Experiment results are shown in the Table~\ref{tab:mt10} and Figure~\ref{fig:mt10}.
        We report the average and standard deviation of the success rate in the evaluation across five random seeds.
        We can observe that FuRL and FuRL-without-image generally outperform the other baselines in most tasks.
        In addition, none of these methods is able to solve the \textit{peg-insert-side} and \textit{pick-place} tasks.
        The main reason is that these two tasks require the agent to master multiple subtasks, \textit{i.e.}, grabbing an item and then moving to a target position, which is highly challenging under a sparse reward setting.
        All the methods struggle on these tasks including those with VLM rewards.
        This is a common weakness within the existing VLM as reward framework and how to go beyond to address this issue is an interesting future direction.

        From Table~\ref{tab:mt10}, we can also observe that the VLM-RMs and VLM-RMs-GB fail to solve any tasks.
        This is not surprising since no task reward is used in VLM-RMs and VLM-RMs-GB,
        and purely the VLM reward is used.
        This exemplifies that it is hard to only rely on the zero-shot VLM rewards
        since there is no guarantee on the reward alignment.
        Moreover, the lower performance of LIV compared to SAC further illustrates the fuzzy reward effect, showing that naively using VLM rewards in online RL can perform worse than the SAC baseline, leading to policies getting stuck in local minima as shown in Figure~\ref{fig:challenges}.
        The better performance of FuRL compared to Relay proves the benefits of reward alignment in mitigating the issue of fuzzy VLM rewards.

        We also evaluate the proposed FuRL on the random-goal MT10 benchmark, where the goal position changes in each trajectory.
        We compare FuRL with the SAC and Relay baselines in the Table~\ref{tab:dynamic}.
        We can observe that FuRL also outperforms other baselines, which indicates that FuRL is able to generalize well with different goal positions.

        \begin{table}[t]
            \caption{\textbf{Experiment on Sparse MT10 with random goals.}}
            % \vspace{-0.05in}
            \label{tab:dynamic}
            % \vskip 0.15in
            \begin{center}
            \resizebox{8.3cm}{!}{
                \begin{tabular}{lccc}
                \toprule
                Environment & SAC & Relay & FuRL  \\
                \midrule
                button-press-topdown-v2 & 16.0 (32.0) & 56.0 (38.3) & 64.0 (32.6) \\
                door-open-v2            & 78.0 (39.2) & 80.0 (30.3) & 96.0 (8.0) \\
                drawer-close-v2         & 100.0 (0.0) & 100.0 (0.0) & 100.0 (0.0) \\
                drawer-open-v2          & 40.0 (49.0) & 50.0 (42.0) & 84.0 (27.3) \\
                pick-place-v2           & 0.0 (0.0) &  0.0 (0.0) &  0.0 (0.0)\\
                peg-insert-side-v2      & 0.0 (0.0) &  0.0 (0.0) &  0.0 (0.0) \\
                push-v2                 & 0.0 (0.0) &  0.0 (0.0) &  6.0 (8.0) \\
                reach-v2                & 100.0 (0.0) & 100.0 (0.0) &  100.0 (0.0) \\
                window-close-v2         & 86.0 (28.0) & 96.0 (4.9) &  100.0 (0.0) \\
                window-open-v2          & 78.0 (39.2) &  92.0 (7.5) & 96.0 (4.9) \\
                \midrule 
                average &  49.8 (7.9) & 57.4 (7.0) & \textbf{64.6 (5.0)} \\
                \bottomrule
                \end{tabular}
                }
            \end{center}
            % \vskip -0.1in
            % \vspace{-0.3in}
        \end{table}

    \subsection{Results with Pixel-based Observations}
\begin{figure}[h]
            \centering
            \begin{overpic}[width=0.7\columnwidth]{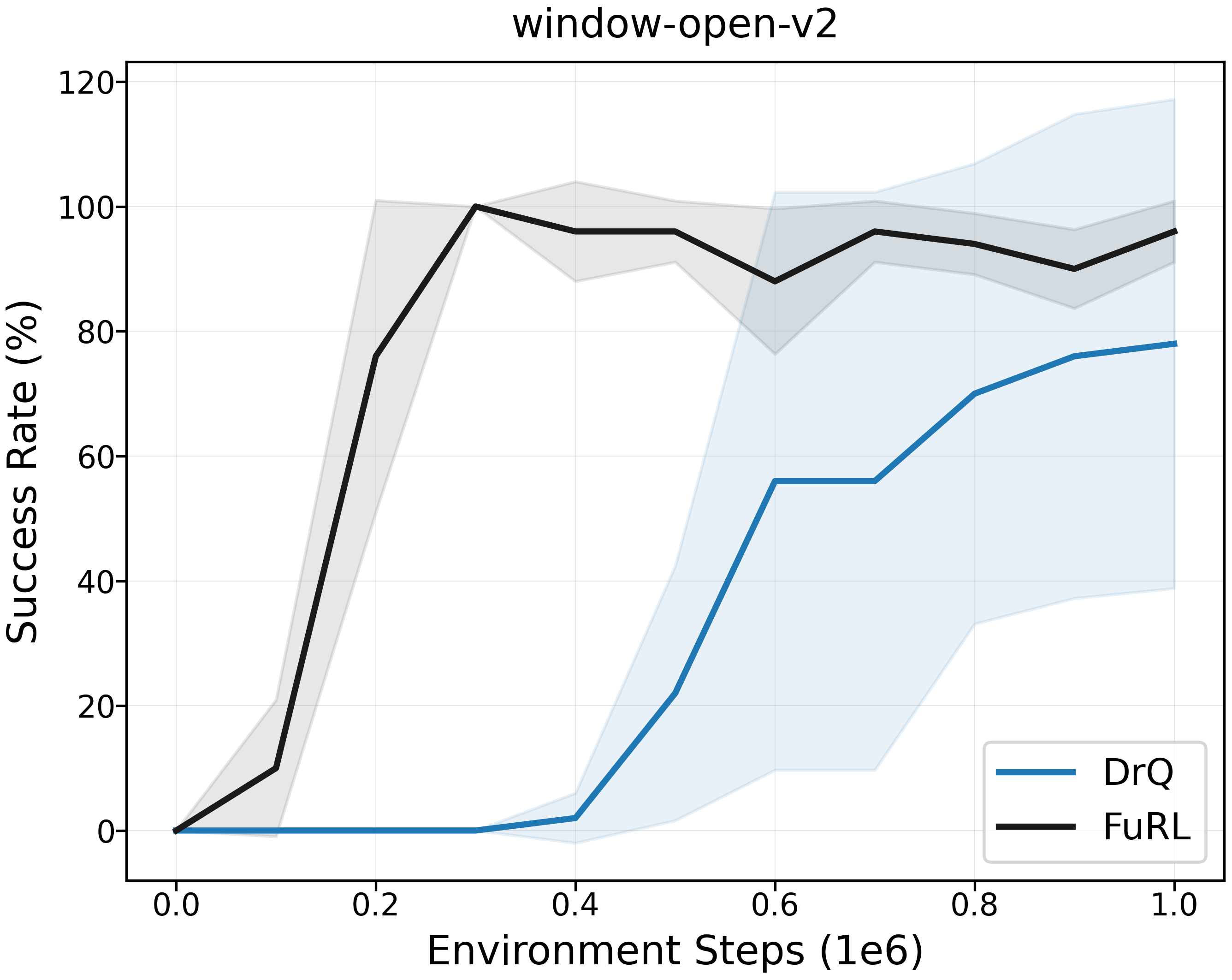}
            \end{overpic}
            \vspace{-0.1in}
            \caption{\textbf{Results on the window-open task with pixel-based observations}: FuRL is also effective with pixel-based inputs.}
            % \vspace{-0.2in}
            \label{fig:pixel}
        \end{figure}

        We further evaluate the proposed method with pixel-based observations.
        More specifically, we adopt the DrQ~\citep{drq, drqv2} as the backbone RL agent.
        We mainly follow the settings from the DrQ to use an image size of 84, stacking frame of 3 and action repeat of 2.
        Figure~\ref{fig:pixel} shows the results on the \textit{window-open} tasks with random goals.
        We can observe that the evaluation curve of FuRL is quite similar to the result in the state-based experiments, which learns much faster than the DrQ baseline.

    \subsection{Ablation Studies}
        We conducted ablation studies to validate the efficacy of the different components in FuRL.
        We used the average success rate and the average Area Under the Curve (AUC) as evaluation metrics.
        To facilitate comparison, we normalized both metrics with respect to the FuRL results.

        \begin{figure}[tb]
            % \vspace{-0.05in}
            \centering
            \begin{overpic}[width=0.9\columnwidth]{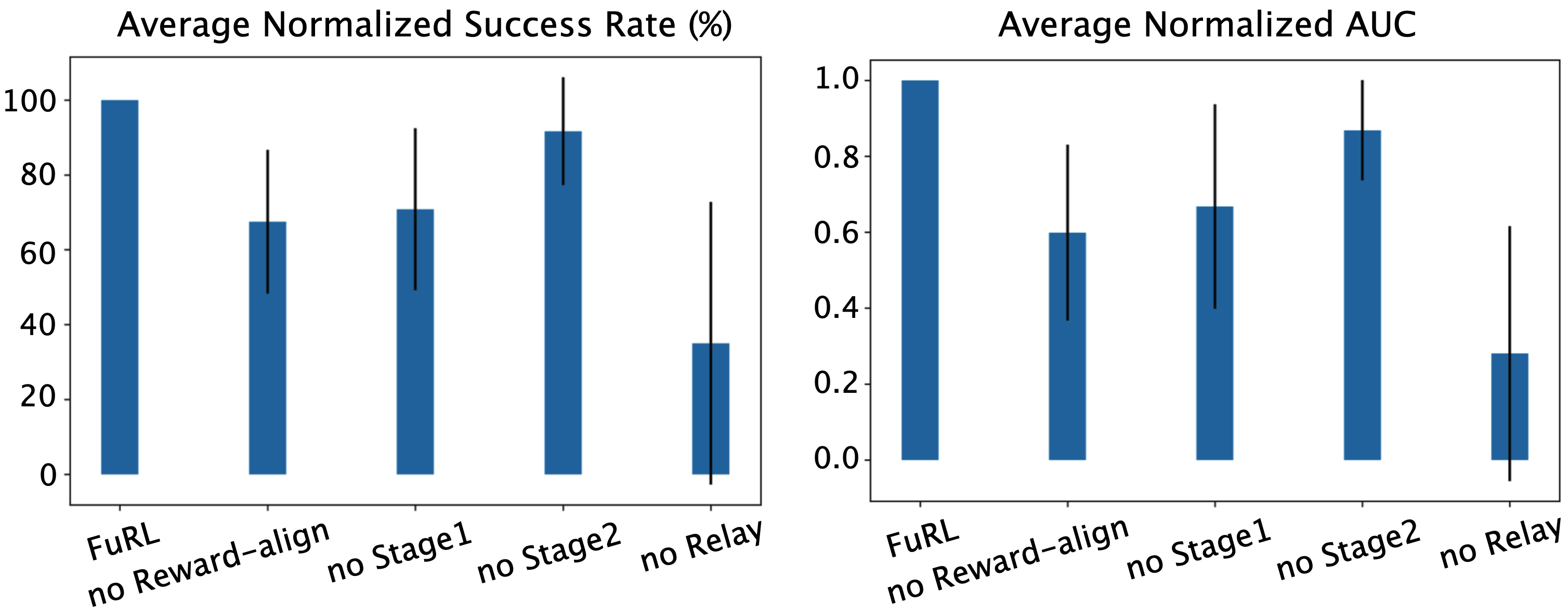}
            \end{overpic}
            % \vspace{-0.15in}
            \caption{\textbf{Ablations.} Reward alignment is important and leads to better performance (\emph{FuRL} v.s. \emph{no Reward-align}). Within the reward alignment part, both Stage 1 (Eqn.~\ref{eq:loss_neg}) and Stage 2 (Eqn.~\ref{eq:reward_alignment}) have contributions. Relay RL is crucial for tasks where the exploration is hard, \textit{i.e.}, the VLM policy is prone to get stuck in local minima due to inaccurate VLM rewards.}
            \vspace{0.05in}
            \label{fig:ablation}
        \end{figure}

        {\textbf{The Impact of Reward Alignment.}}
            We first validate the effectiveness of the proposed reward alignment loss functions in Figure~\ref{fig:ablation}.
            The \emph{no Reward-align} baseline is a variant of FuRL without reward alignment.
            We can observe that \emph{no Reward-align} baseline consistently underperforms FuRL, which demonstrates the efficacy of the reward alignment component.
            Furthermore, we refer \textit{no Stage1} to fine-tuning with only Eqn.~\ref{eq:reward_alignment}, and we refer \textit{no Stage2} to fine-tuning with only Eqn.~\ref{eq:loss_neg}.
            We can observe that removing any of Stage1 or Stage2 will lead to performance degradation, \textit{i.e.}, slower convergence and (or) higher variance.
            This shows that both loss functions in the reward alignment component are effective, where the Eqn.~\ref{eq:reward_alignment} helps to mitigate the representation misalignment issue, and Eqn.~\ref{eq:loss_neg} helps to find the first successful trajectory earlier.

        {\textbf{The Impact of Relay RL.}}
            From Figure~\ref{fig:ablation}, we can also observe that the relay RL plays a crucial role in FuRL.
            Without relay RL, the agent completely failed in the some tasks, \textit{i.e.}, \textit{button-press-topdown} and \textit{door-open} tasks.
            This is because the reward alignment loss function Eqn.~\ref{eq:reward_alignment} is triggered after we collected the first successful trajectory.
            The relay RL addresses the challenge of getting stuck in local minima when we only have negative samples.

        {\textbf{{The Impact of VLM Embedding.}}
            Figure~\ref{fig:no_vlm_embedding} shows the results of training a FuRL agent without the pre-trained VLM-representation.
            We can observe that the variant without VLM-representation performs worse than FuRL and performs better than the SAC baseline. This verifies the benefits of using the VLM model and also the effectiveness of the proposed reward alignment objective.

        \begin{figure}[h]
            % \vspace{-0.05in}
            \centering
            \begin{overpic}[width=0.95\columnwidth]{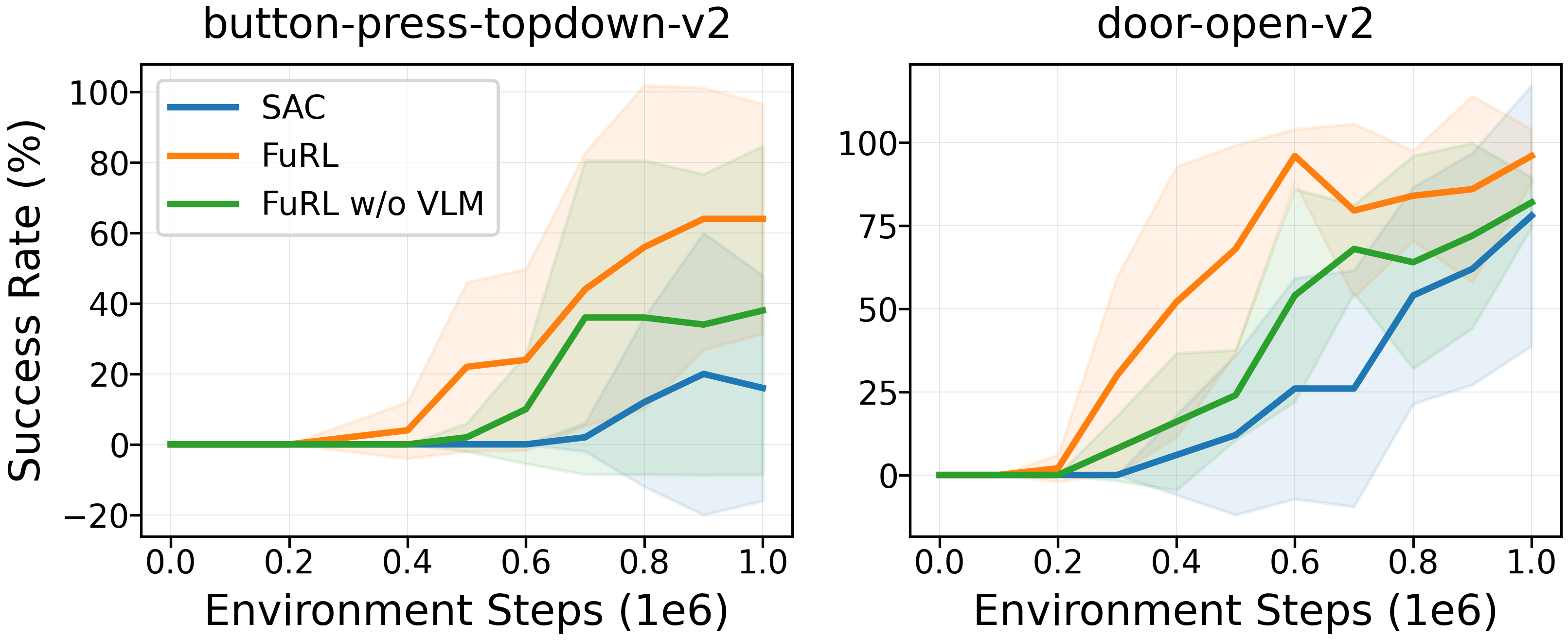}
            \end{overpic}
            % \vspace{-0.1in}
            \caption{\textbf{Impact of VLM-representation.}}
            % \vspace{-0.1in}
            \label{fig:no_vlm_embedding}
        \end{figure}
        
        {\textbf{Using CLIP as the VLM model.}}
            In the previous experiments, we instantiate the VLM model as a pre-trained LIV model~\citep{liv}.
            In this subsection, we evaluate the proposed FuRL by instantiating the VLM model as a pre-trained CLIP~\citep{clip}.
            Results on MT10 with fixed-goal are shown in Table~\ref{tab:clip}.
            The CLIP baseline is a SAC agent that learns with task reward and dense CLIP reward.
            We can observe that the CLIP-based FuRL also outperforms the CLIP baseline, which shows that the proposed method can generalize to different VLM base models.
    
            \begin{table}[t]
                % \vspace{-0.1in}
                \caption{\textbf{Experiment of using CLIP as the VLM.}}
                \label{tab:clip}
                \vspace{0.05in}
                \footnotesize
                \begin{center}
                    \begin{tabular}{lcc}
                    \toprule
                    Environment & CLIP & CLIP-FuRL  \\
                    \midrule
                    button-press-topdown-v2 & 0 (0) & 80 (40) \\
                    door-open-v2            & 60 (49) & 100 (0) \\
                    drawer-close-v2         & 100 (0) & 100 (0) \\
                    drawer-open-v2          & 0 (0) & 80 (40) \\
                    peg-insert-side-v2      & 0 (0) & 0 (0) \\
                    pick-place-v2           & 0 (0) & 0 (0) \\
                    push-v2                 & 0 (0) & 60 (49) \\
                    reach-v2                & 80 (40) & 100 (0) \\
                    window-close-v2         & 60 (49) & 100 (0) \\
                    window-open-v2          & 80 (40) & 80 (40) \\
                    \midrule 
                    average & 38 (9.8) & \textbf{70 (6.3)} \\
                    \bottomrule
                    \end{tabular}
                \end{center}
                \vspace{-0.1in}
            \end{table}

        \begin{figure}[h]
            \centering
            \begin{overpic}[width=0.9\columnwidth]{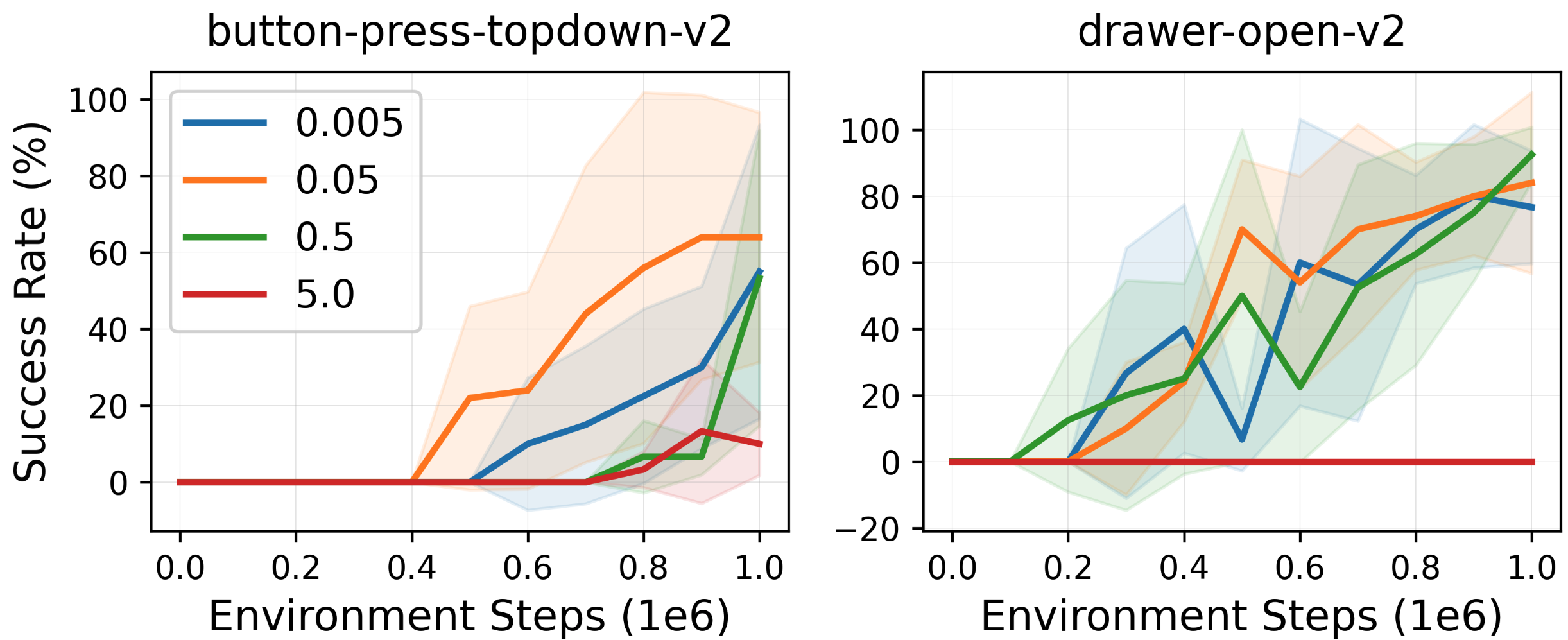}
            \end{overpic}
            \caption{\textbf{Impact of parameter $\rho$}.}
            \label{fig:rhos}
        \end{figure}
        {\textbf{{The Impact of $\rho$.}}
            Figure~\ref{fig:rhos} shows the results of using different $\rho$ values on the \textit{drawer-open} and \textit{button-press-topdown} tasks with random goals.
            We can observe that a large value usually performs poorly, where the inaccurate VLM reward distracts the agent from learning from the task reward information.
            On the other hand, a small value might lead to slow learning or larger variance. 

    \subsection{Visualization of VLM Rewards}
        Figure~\ref{fig:vlm_repr2} shows the VLM rewards before (LIV) and after alignment (FuRL) for the trajectory shown in Figure~\ref{fig:vlm_repr}.
        Compared with the pre-trained LIV reward curve, the FuRL VLM reward curve generally has a larger value when it is closer to the goal.
        The reason that the reward does not reach 1 after alignment is because a ranking loss is used in Eqn.~\ref{eq:reward_alignment}, which focuses on the relative ranking instead of absolute reward values. 
        The accurate relative trend in the aligned reward from FuRL is already effective in aiding the RL agent in exploration and learning.

        \begin{figure}[h]
            \centering
            \vspace{0.05in}
            \begin{overpic}[width=1.0\columnwidth]{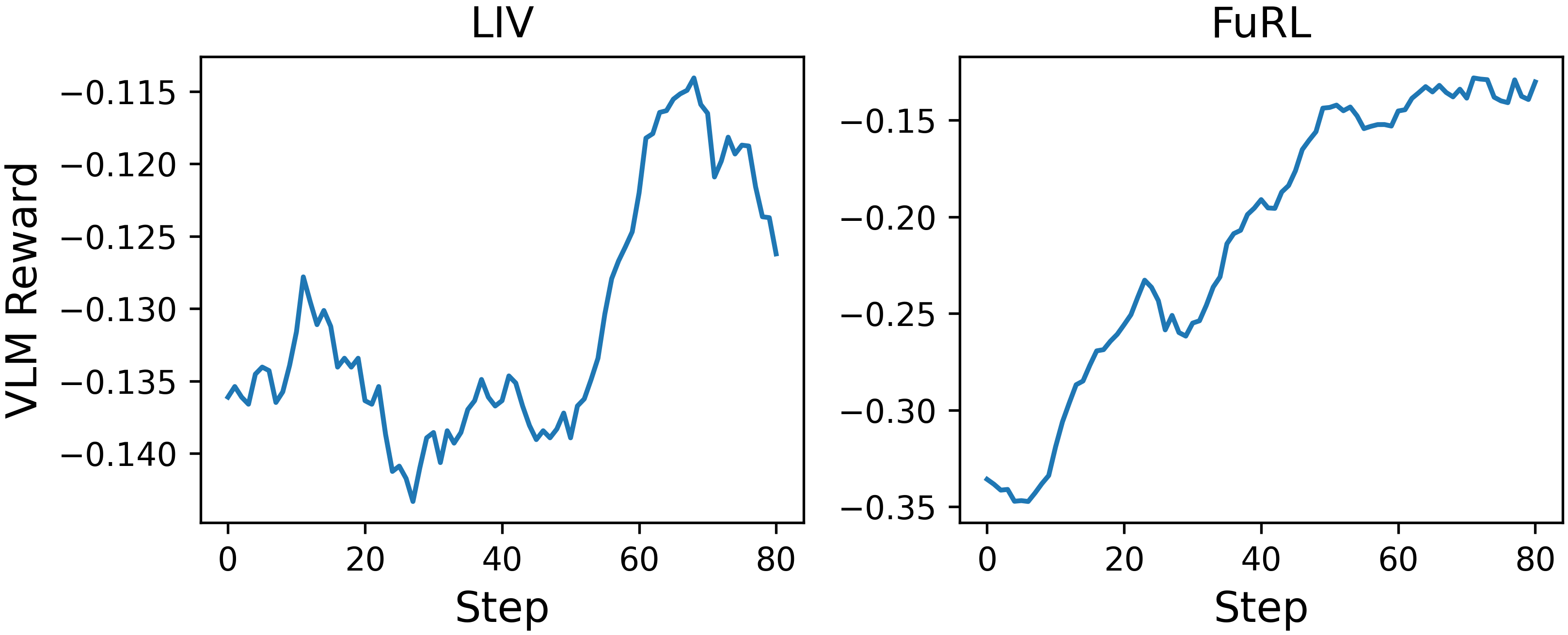}
            \end{overpic}
            \vspace{-0.25in}
            \caption{\textbf{VLM reward curves} before (left) and after (right) fine-tuning, along the same expert trajectory. Some frames along the trajectory are shown in Figure~\ref{fig:vlm_repr}. FuRL reward has a larger value when it is closer to the goal.}
            \vspace{-0.2in}
            \label{fig:vlm_repr2}
        \end{figure}
\section{Conclusion and Future Work}
    \label{sec:conclusion}
    In this work, we highlighted the fuzzy reward issue in applying VLM reward in online RL and
    further proposed the FuRL approach,  which mitigated the fuzzy reward issue with reward alignment and relay RL  collaboratively. Various experiments demonstrated the effectiveness of the proposed method.
    In the current work,
    we have demonstrated the approach on a number of tasks, with the reward alignment networks trained with data from a single task. 
    An interesting future direction is to train the reward alignment module across multiple task jointly.
    Another point is that we have used simple language descriptions following~\citet{rocamonde2023vision, liv}. Applying the proposed approach to more complex compositional language instructions is also an interesting direction to pursue in the future.
    From a broader perspective, the hallucination issue of VLMs and large language models (LLMs)~\citep{li2023evaluating, chakraborty2024hallucination} has greatly impacted their applicability in downstream tasks. It is an interesting future work to generalize some of the ideas in this work together with other techniques such as adversarial learning~\citep{adv_fea, feature_scatter, yao2023llm} to
    a broader context with pre-trained foundations models.

\section*{Impact Statement}
    Since our work is a combination of VLM and RL, one potential negative societal impact could be the improper language instructions.
    When we apply the proposed method to real-world applications, the usage of some language instructions might lead to dangerous behaviours.
    To mitigate this issue, we could adopt some rule-based keyword blacklists to filter dangerous language instructions, or we could further fine-tune the trained policy to learn some safety knowledge.

\vspace{-0.05in}
\section*{Acknowledgements}
We would like to thank the anonymous reviewers 
    for their efforts in reviewing our work and their constructive comments that help to further improve the quality of this paper.
\bibliography{reference}
\bibliographystyle{icml2024}
\newpage
\appendix
\onecolumn

\section{Limitations, Potential Broader Impact, Ethical Aspects and Societal Impacts}
    In this work, we mainly focus on illustrating the \textit{Fuzzy-reward} effect and why it could be problematic to directly apply the pre-trained VLM rewards in online RL tasks.
    One limitation is that our current experiments are all in simulated environments.
    We plan to validate the effectiveness of the proposed method in real-world robotics in future work.
    Another limitation is that we maintained a second policy for Relay RL to escape the local minima during the online exploration.
    Though we turn off the Relay RL when we collect some successful trajectories, the extra policy still increases the computation complexity.
    An interesting future direction is to replace the replay policy by some lightweight exploration intrinsic reward to mitigate the issue of getting stuck in local minima.

    Since our work is a combination of VLM and RL, one potential negative societal impact could be the improper language instructions.
    When we apply the proposed method to real-world applications, the usage of some language instructions might lead to dangerous behaviours.
    To mitigate this issue, we could adopt some rule-based keyword blacklists to filter dangerous language instructions, or we could further fine-tune the trained policy to learn some safety knowledge.

\section{Experimental Setup}
    \label{appendix:exp_setup}

    \subsection{Implementation Details}
        In the experiment, we re-implement the SAC~\citep{sac} and DrQ~\citep{drq} baseline RL agents in JAX~\citep{jax}.
        For the VLM model, we use the provided PyTorch code~\citep{imambi2021pytorch} and checkpoint for both of LIV and CLIP from the official LIV codebase\footnote{https://github.com/penn-pal-lab/LIV}.
        In the experiments, we use the latest Meta-world environment \footnote{https://github.com/Farama-Foundation/Metaworld}.
        For the other main softwares, we use the following versions:

        {
        \begin{itemize}
            \item Python 3.9
            \item jax 0.4.16
            \item jaxlib-0.4.16+cuda12.cudnn89-cp39
            \item flax 0.7.4
            \item gymnasium 0.29.1
            \item imageio 2.33.1
            \item optax 0.1.7
            \item torch 2.1.2
            \item torchvision 0.16.2
            \item numpy 1.26.2
        \end{itemize}
        }

        \begin{figure}[h]
            \vskip 0.2in
            \begin{center}
            \centerline{\includegraphics[width=16cm]{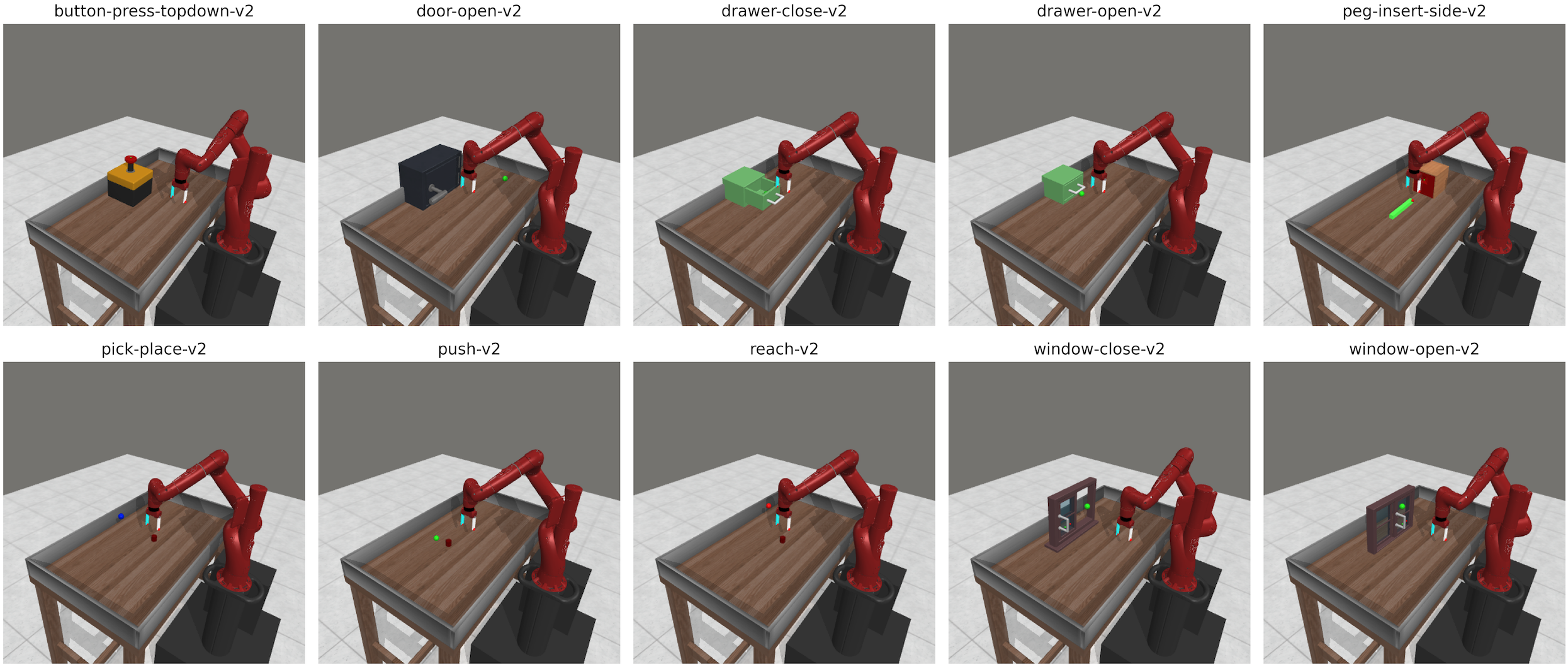}}
            \caption{Meta-world MT10 benchmark tasks.}
            \label{fig:mt_envs}
            \end{center}
            \vskip -0.2in
        \end{figure}
    \subsection{Meta-world MT10 Benchmark}
        In the experiments, we use a constant reward shaping $r^{\rm task}$ - 1 for the sparse task reward as in some previous work~\citep{iql}.
        For the task description for each environment, we followed the setting from the CARE~\citep{sodhani2021multi}.

        In the experiments of Table~\ref{tab:mt10}, the goal is hidden and fixed for each random seed.
        Since we always use the mean action of the SAC policy in the evaluation and the Meta-world environment is deterministic, the evaluation success rate for each task is either 1 or 0 for each random seed.
        For the results of random goal setting in Table~\ref{tab:dynamic}, we report the average evaluation success rate over ten trajectories.
        For the average results across all tasks in the Table~\ref{tab:mt10}, Table~\ref{tab:dynamic}, and Table~\ref{tab:clip}, we first group experiments for one algorithm into five runs across ten tasks, and get an ten-task average success rate, and compute the standard deviation based on the five numbers.
        For the goal image, we simply use a fixed goal image for both fixed goal and random goal tasks.
        The main idea of the goal image is to provide some useful information to distinguish two negative samples.
        Therefore, we adopt this simple setting without the loss of generality.

        \begin{table}[t]
            \caption{Text instruction for each environment in the experiments.}
            \label{tab:language_goal}
            \vskip 0.15in
            \begin{center}
                \resizebox{9cm}{!}{
                \begin{tabular}{lc}
                \toprule
                Environment             & Text instruction \\
                \midrule
                button-press-topdown-v2 & Press a button from the top. \\
                door-open-v2            & Open a door with a revolving joint. \\
                drawer-close-v2         & Push and close a drawer. \\
                drawer-open-v2          & Open a drawer. \\
                peg-insert-side-v2      & Insert a peg sideways. \\
                pick-place-v2           & Pick and place a puck to a goal. \\
                push-v2                 & Push the puck to a goal. \\
                reach-v2                & Reach a goal position. \\
                window-close-v2         & Push and close a window. \\
                window-open-v2          & Push and open a window. \\
                \bottomrule
                \end{tabular}
                }
            \end{center}
            \vspace{-0.1in}
        \end{table}

        \begin{table}[h]
            \caption{Summarization of hyper-parameters.}
            \label{tab:parameters}
            % \vskip 0.15in
            \begin{center}
            \resizebox{7.0cm}{!}{
                \begin{tabular}{lc}
                \toprule
                Parameter                & Value \\
                \midrule
                Total environment step   & 1e6   \\
                Adam learning rate       & 1e-4  \\
                Batch size               & 256   \\
                Camera Id                & 2     \\
                VLM reward weight $\rho$ & 0.05  \\
                Target network $\tau$    & 0.01  \\
                Discount factor $\gamma$ & 0.99  \\ \midrule 
                FuRL language projection network & (256, 64) \\
                FuRL image projection network & (256, 64) \\ 
                FuRL window size $k$     & 10 \\
                FuRL reward margin       & 0.1 \\
                FuRL L2 distance margin  & 0.2 \\ \midrule
                SAC buffer size          & 1e6 \\
                SAC actor network        & (256, 256) \\
                SAC critic network       & (256, 256) \\
                SAC target entropy       & -$\vert \mathcal A \vert$/2 \\ \midrule
                DrQ buffer size          & 2e5 \\
                DrQ action repeat       & 2 \\
                DrQ frame stack          & 3 \\
                DrQ image size           & (84, 84, 3) \\
                DrQ embedding dimension  & 50 \\
                DrQ CNN features         & (32, 32, 32, 32) \\
                DrQ CNN kernels          & (3, 3, 3, 3) \\
                DrQ CNN strides          & (2, 1, 1, 1) \\
                DrQ CNN padding          & VALID \\
                DrQ actor network        & (256, 256, 256) \\
                DrQ critic network       & (256, 256, 256) \\
                \bottomrule
                \end{tabular}
                }
            \end{center}
        \end{table}

        \begin{figure}[h]
            \vskip 0.2in
            \begin{center}
            \centerline{\includegraphics[width=16cm]{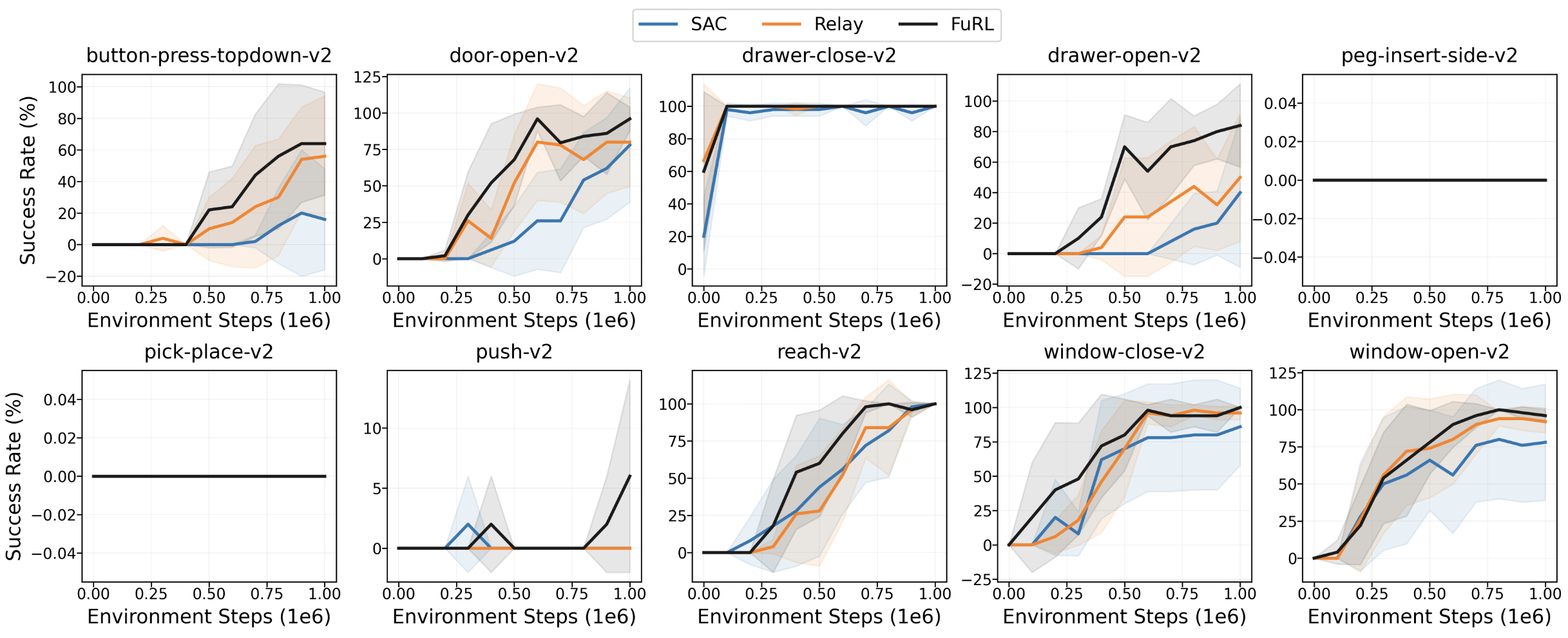}}
            \caption{\textbf{Evaluation curves for tasks with random goal.}}
            \label{fig:random_goal}
            \end{center}
            \vskip -0.2in
        \end{figure}

    \subsection{Computation Complexity}
        We run our experiments on a workstation with NVIDIA GeForce RTX 3090 GPU and a 12th Gen Intel(R) Core(TM) i9-12900KF CPU.
        The average wall-clock running time for the FuRL on the state-based experiment and pixel-based experiment are 3 hours and 6.5 hours, respectively.

    \subsection{Parameter Settings}
        Some key parameters are summarized in the Table~\ref{tab:parameters}.
        We mainly followed the parameter settings from some prior work~\citep{vip,liv,drq,drqv2}.
        For the relay steps, we select $Ts$ to be a set of four discrete values [50, 100, 150, 200] out of simplicity.
        Some other choices, \textit{i.e.}, using a uniform distribution U[50, 250], are also acceptable.
        Since the main focus of this work is to illustrate the Fuzzy reward issue and showcase the effectiveness of the proposed method, therefore, we did not tune much of these hyper-parameters.
        It is likely to achieve better performances with further parameter tuning.

\section{Additional Experiment Results}

    \subsection{Experiment on Sparse MT10 with random goals}

        \begin{figure}[htb]
            \vskip 0.2in
            \begin{center}
            \centerline{\includegraphics[width=16cm]{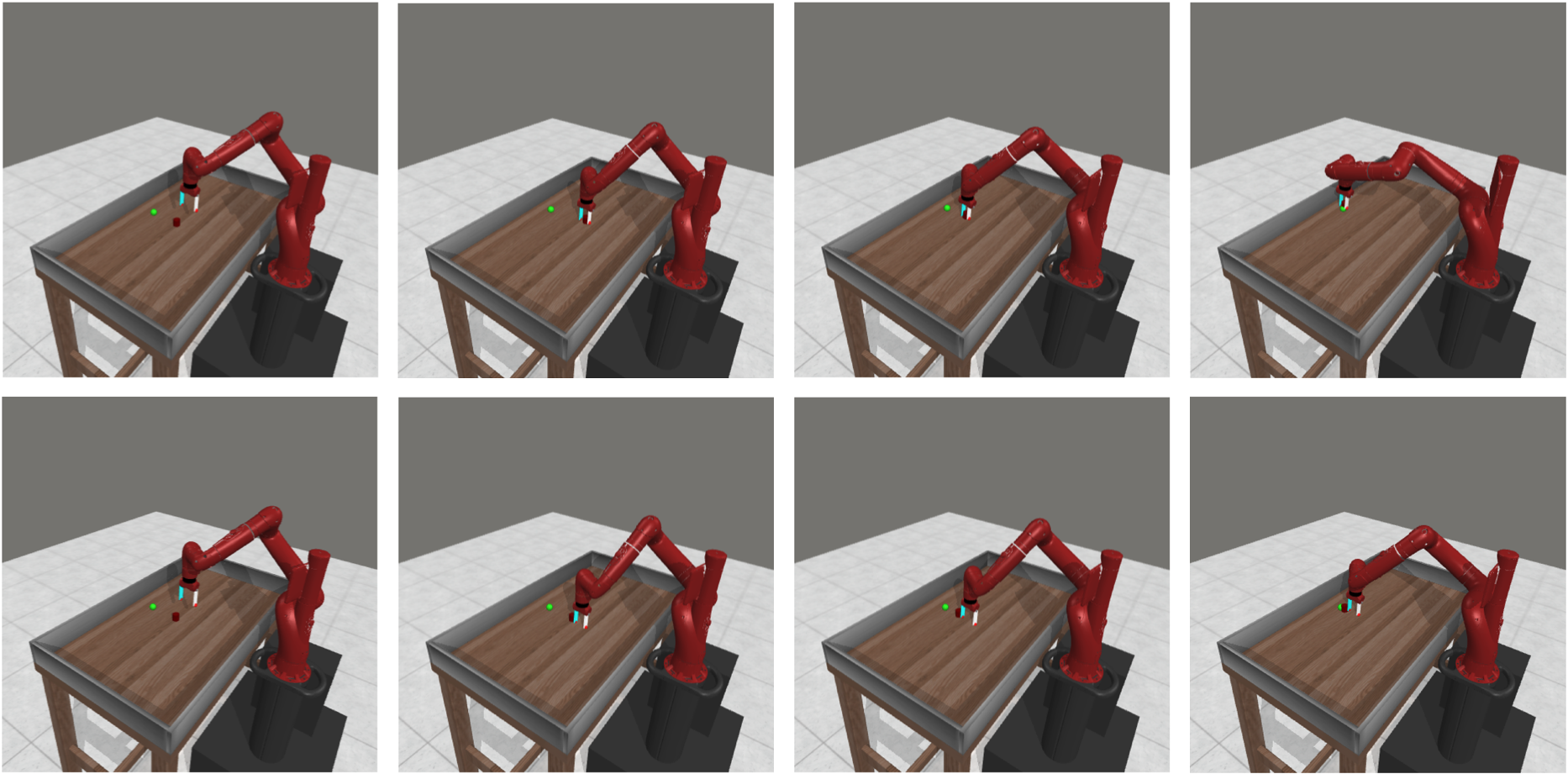}}
            \caption{\textbf{Comparison of the oracle policy and the FuRL policy.} The oracle policy first grasps the red cylinder and then move to the green goal position (top row). On the other hand, the learned FuRL policy simply pushes the cylinder to the goal position (bottom row).}
            \label{fig:push_oracle_furl}
            \end{center}
            \vskip -0.2in
        \end{figure}

        Figure~\ref{fig:random_goal} shows the results of SAC, Relay and FuRL on Sparse MT10 with random goals.
        Similar to the conclusion as in Figure~\ref{fig:mt10}, FuRL generally outperforms the SAC and Relay baselines.
        Compared with the fixed goal setting, we can observe that FuRL generally achieved similar performances except for the \textit{push-v2} task. Figure~\ref{fig:push_oracle_furl} shows two successful trajectories for the oracle policy and the FuRL policy. We can observe that the oracle policy first grasps the red cylinder and then moves to the green goal position. On the other hand, the FuRL policy didn't learn how grasp the cylinder and just simply pushes the cylinder to the goal position. Compared with the oracle policy, the FuRL policy is less robust without grasping the object, especially when the goal position is far away from the initial object position and the cylinder falls down and starts rolling. Moreover, another challenge in the \textit{push-v2} task is that the arm sometimes blocks the goal and (or) the cylinder in the camera due to their small sizes. This makes the random goal setting much more difficult because the current VLM reward only relies on the input image, and an image without the goal provides less informative VLM reward.

    \subsection{The impact of language instructions}
        In the previous experiments, we all used the same language instructions.
        Here, we try to investigate how the input language instruction affects the final performance.
        We compare three different language instructions for the \textit{button-press-topdown} and \textit{drawer-open} tasks.
        The first language instruction is a dummy input text of ``None''.
        The other two language instructions are summarized in Table~\ref{tab:long_language_goal}.
        From Figure~\ref{fig:text}, we can observe that using a more detailed language instruction can help to learn faster, while using a meaningless language instruction usually performs poorly.
        These results indicate that a more detailed language instruction sometimes could provide more accurate VLM rewards that help the agent to find the first successful trajectory earlier, and a misleading language instruction could provide inaccurate VLM rewards, which are more likely to trap the agent in local minima.
        Since the focus of this work is to illustrate the issue of Fuzzy VLM rewards and how to address this issue, we leave the exploration of how to generate better text instructions for future work.

        \begin{figure}[htb]
            \vskip 0.2in
            \begin{center}
            \centerline{\includegraphics[width=16cm]{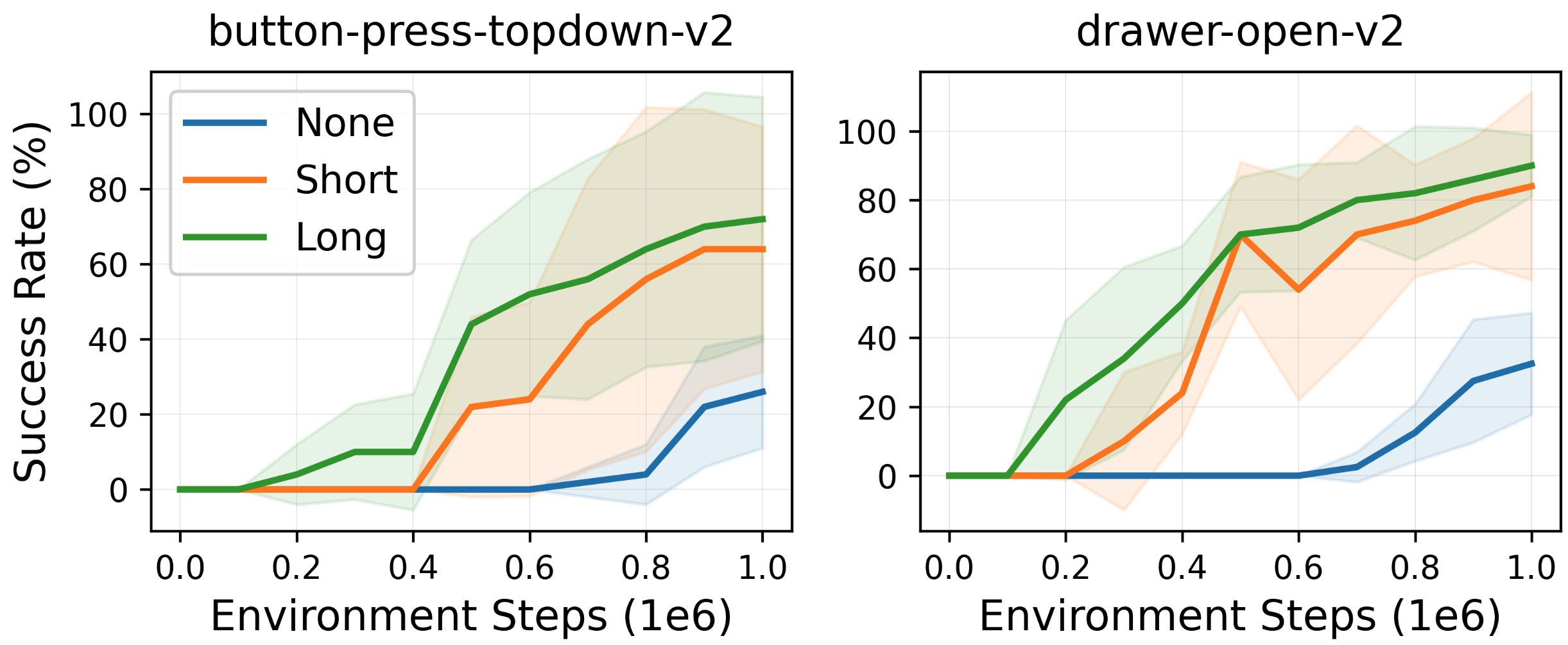}}
            \caption{\textbf{Ablation for different language instructions.}}
            \label{fig:text}
            \end{center}
            \vskip -0.2in
        \end{figure}

        \begin{table}[htb]
            \caption{Different text instructions.}
            \label{tab:long_language_goal}
            \vskip 0.15in
            \begin{center}
                \resizebox{13.5cm}{!}{
                \begin{tabular}{lcc}
                \toprule
                Environment             & Length & Text instruction \\
                \midrule
                \multirow{2}{*}{button-press-topdown-v2 } & Short & Press a button from the top. \\ 
                    & Long & Move close to the orange box and press down the red button.  \\ \midrule
                \multirow{2}{*}{drawer-open-v2}          & Short & Open a drawer. \\ 
                    & Long & Grab the white handle and open the green drawer. \\
                \bottomrule
                \end{tabular}
                }
            \end{center}
            \vspace{-0.1in}
            % \vskip -0.1in
        \end{table}

    \subsection{Visualization of Inaccurate VLM Reward}
        In Figure~\ref{fig:oracle_suboptimal_random_traj}, we plot three trajectories for an oracle policy $\pi_o$, a sub-optimal policy $\pi_s$ and a random policy $\pi_r$.
        The sub-optimal policy is a policy been trained for small number of steps (1e5), which can move the arm around but is not able to complete the task yet (task success rate is 0\%).

        In Figure~\ref{fig:oracle_suboptimal_random}, we plot the VLM-rewards at different steps for the oracle policy $\pi_o$, random policy $\pi_r$ and sub-optimal policy $\pi_s$ for 50 trajectories, showing both individual curves as well as mean-std curves.
        We can observe that the random policy $\pi_r$ is not very informative in this case since a random policy mostly causes the robot arm to jitter around its initial position and can barely move the arm.
        In addition, we can observe that the VLM-reward is very noisy.
        For example, there is a significant overlap between the curves from oracle and sub-optimal policies, although their level of expertise is drastically different, meaning the set of states covered by them are very different.
        Moreover, for the sub-optimal trajectories, there are cases where the VLM assigns very high reward to some sub-optimal states (e.g. around step 40, and step 80), with values comparable to (sometime even higher than) the VLM-reward for expert trajectory's success state (expert's trajectory around step 80).
        All these illustrate the fuzzy VLM-reward issue. Learning using this type of reward could mislead or trap RL agent in a local minimum and leading to undesired behaviors as shown in Figure~\ref{fig:challenges} of the main paper.
        
        \begin{figure*}[htb]
            \vspace{-0.05in}
            \centering
            \begin{overpic}[width=0.95\columnwidth]{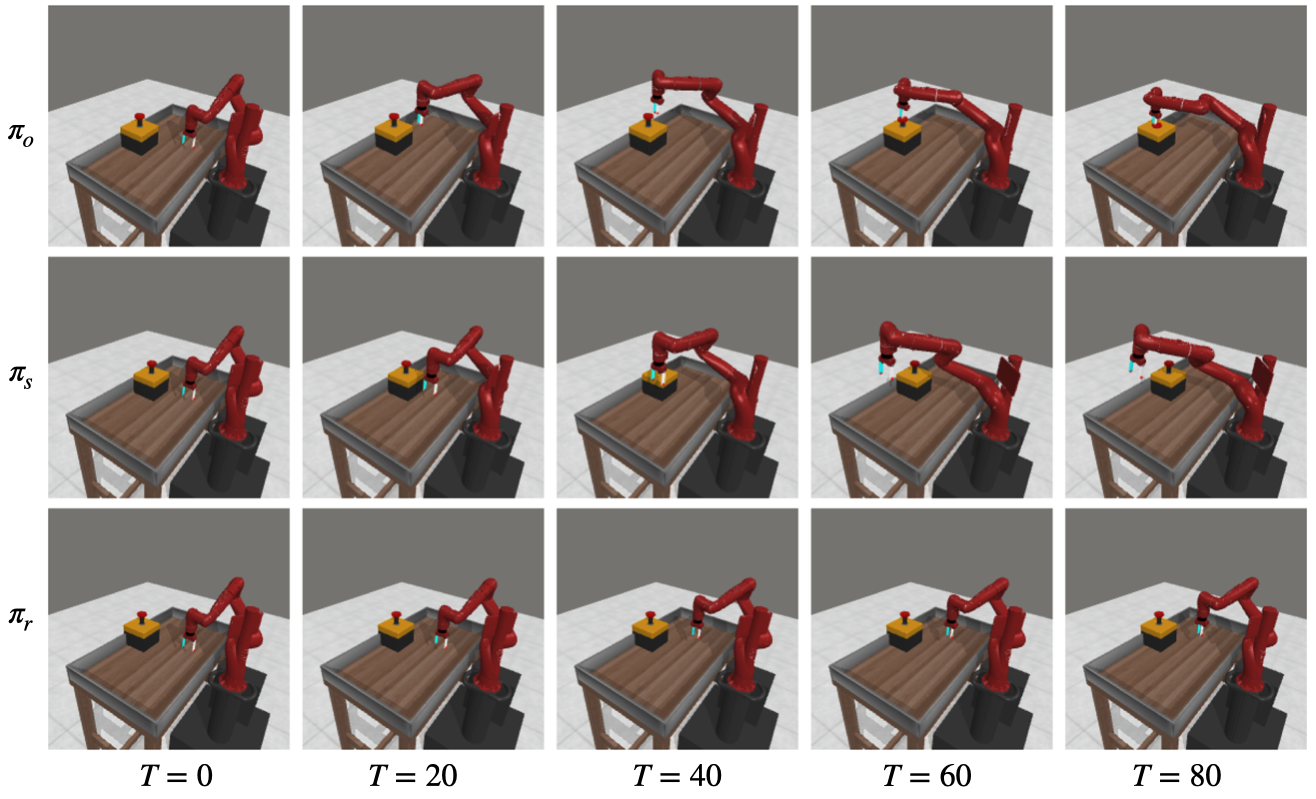}
            \end{overpic}
            \vspace{-0.15in}
            \caption{\textbf{More trajectory visualizations:} (top) an oracle policy $\pi_o$, (middle) a sub-optimal policy $\pi_s$, and (bottom) a random policy $\pi_r$.}
            % \vspace{-0.15in}
            \label{fig:oracle_suboptimal_random_traj}
        \end{figure*}

        \begin{figure*}[htb]
            \vspace{-0.05in}
            \centering
            \begin{overpic}[width=0.95\columnwidth]{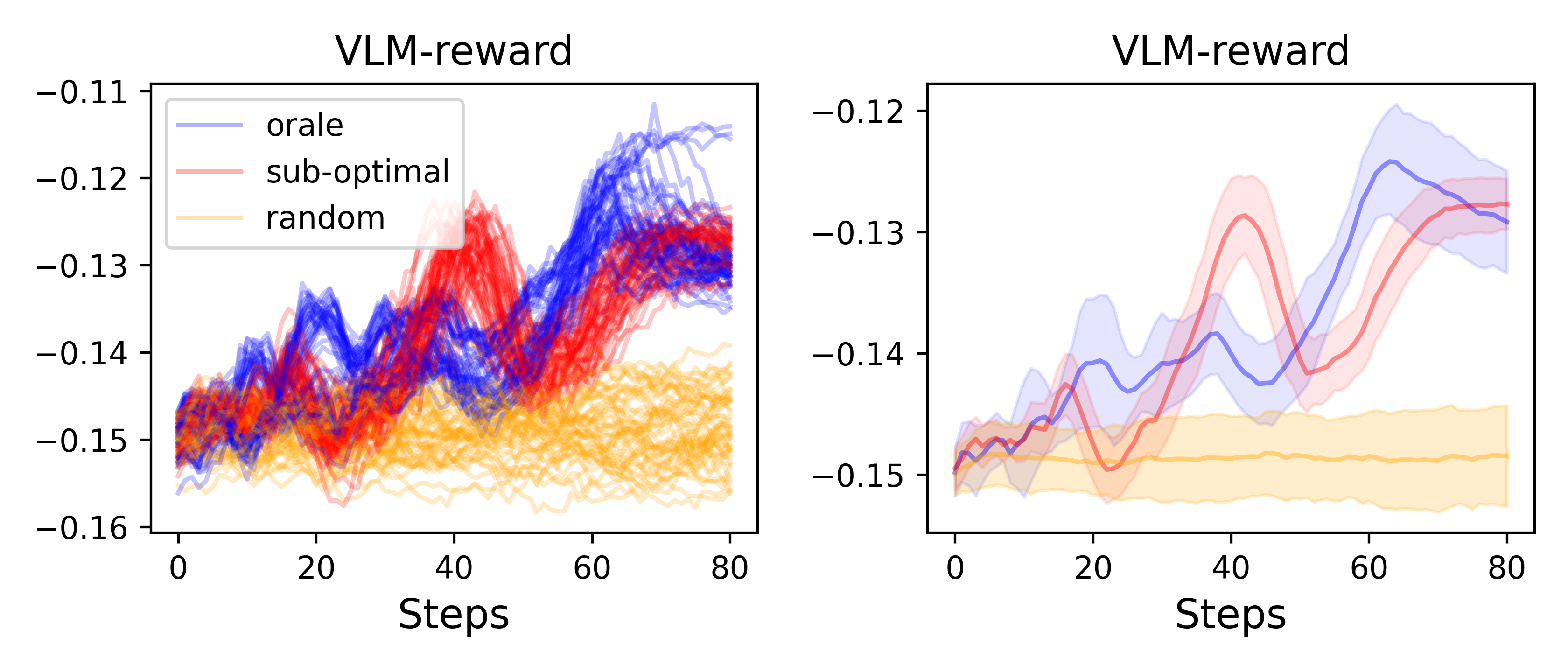}
            \end{overpic}
            \vspace{-0.15in}
            \caption{\textbf{Visualization of noisy VLM-rewards:} there is a significant overlap between the curves from oracle and sub-optimal policies.}
            \vspace{-0.15in}
            \label{fig:oracle_suboptimal_random}
        \end{figure*}

\end{document}